\documentclass[a4paper]{report}

\usepackage{graphicx}
\usepackage{color}
\usepackage[center]{caption}
\usepackage{subcaption}
\usepackage{amsmath}
\usepackage{amssymb}
\usepackage{rotating}
\usepackage{listings}
\usepackage{multirow}
\usepackage{enumitem}
\usepackage{epstopdf}
\usepackage{multirow}
\usepackage{amsbsy}

\usepackage{chngpage}

\usepackage{fancyhdr}
\usepackage{natbib}
\bibpunct{[}{]}{,}{n}{}{}

\usepackage{lettrine}

\usepackage[pdfborder={0 0 0}]{hyperref}

\graphicspath{{./graphics/}}

\begin{document}
\begin{titlepage}
\begin{center}

\includegraphics[width=0.75\textwidth]{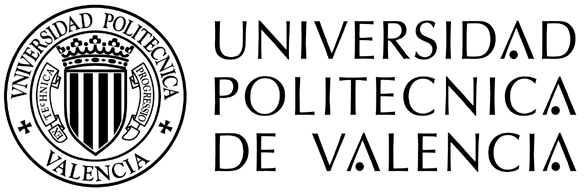}\\[1.5cm]

\textsc{\Large Master in Computer Engineering}\\[0.7cm]

\rule{\textwidth}{1.6pt}\vspace*{-\baselineskip}\vspace*{3.2pt}
\rule{\textwidth}{0.4pt}\\[\baselineskip]
{ \Large \bfseries Master Thesis }\\[0.4cm]
{ \Huge \bfseries Measurements of collective \\[10pt] machine intelligence }\\[0.4cm]
\rule{\textwidth}{0.4pt}\vspace*{-\baselineskip}\vspace*{4.4pt}
\rule{\textwidth}{1.6pt}\\[0.7cm]

\begin{minipage}[t]{0.49\textwidth}
\begin{flushleft} \large
\emph{Student:}\\
Michel \textsc{Halmes} \\
Universit\'e Libre de Bruxelles
\end{flushleft}
\end{minipage}
\begin{minipage}[t]{0.49\textwidth}
\begin{flushright} \large
\emph{Supervisor:} \\
Prof. Jos\'e  \textsc{Hern\'andez-Orallo}  
\end{flushright}
\end{minipage}\\[1.5cm] 

\includegraphics[width=0.75\textwidth]{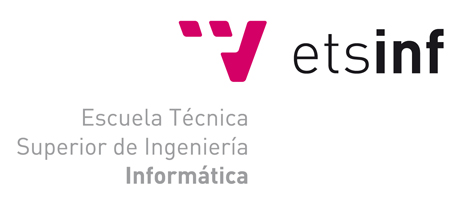}\\
\vfill

{\large \today}

\end{center}
\end{titlepage}

\begin{center}
\begin{minipage}[t]{0.7\textwidth}
\chapter*{Acknowledgment}
\addcontentsline{toc}{chapter}{Acknowledgment}
\mdseries

I would like to thank my supervisor Prof. Jose  Hernandez-Orallo for having supported my master thesis during the whole year. I thank you for the time you considered to this project and for your constructive comments which directed my work.
\newline \hfill \newline 
Finally, I would also like to thank the many people which made out of my Erasmus year in Valencia an unforgettable time.

\end{minipage}
\end{center}
 
\newpage
\chapter*{Abstract}
\addcontentsline{toc}{chapter}{Abstract}

\lettrine{I}{ntelligence} is a fairly intuitive concept of our everyday life. As usually acknowledged in psychometrics, ``intelligence is the ability measured by intelligence tests''. However, defining what exactly intelligence tests should measure is less obvious. During the last decade, computer scientists have attempted to provide a formal definition of intelligence. There seems now to be the  tendency that intelligence should make reference to the formalism provided by the field of algorithmic information theory. Yet, a consensus is far from being reached.

Independent from the  still ongoing research in measuring individual intelligence, we anticipate and provide a framework for measuring collective intelligence. Collective intelligence refers to the idea that several individuals can collaborate in order to achieve high levels of intelligence. We present thus some ideas on how the intelligence of a group can be measured and simulate such tests. We will however focus here on groups of artificial intelligence agents (i.e., machines). We will explore how a group of agents is able to choose the appropriate problem and to specialize for a variety of tasks. This is a feature which is an important contributor to the increase of intelligence in a  group (apart from the addition of more agents and the improvement due to common decision making).

Our results reveal some interesting results about how (collective) intelligence can be modeled, about how collective intelligence tests can be designed and about the underlying dynamics of collective intelligence. As it will be useful for our simulations, we provide also some improvements of the threshold allocation model originally used in the area of swarm intelligence but further generalized here.

\vfill

{\bf Keywords:} collective intelligence, machine intelligence tests, task allocation models, problem specialization, swarm intelligence, universal psychometrics, joint decision making, multi-task evaluation

\newpage
\tableofcontents

\newpage

\pagestyle{fancy}

\chapter{Goal statement and overview}
\label{ch:goal}

There have been some works which have studied the contribution of collective decisions making (e.g., multi-classifier systems and decision theory) and some other works which have focused on optimal allocation (e.g., swarm intelligence and resource allocation). In this work, we analyze both things together, as they are contributors to the observed increase in collective intelligence in many real systems (e.g., brains, societies, biology). Thus, we will explore group performance according to joint decision making, task allocation, several degrees of intelligence (or ability) and agent specialization.

\section{Goal statement}
\label{sec:goalStmt}

The goal of this master thesis is to experiment with collective machine intelligence abstraction models, so as to observe some interesting phenomena in the intellectual capabilities of several, collaborating machines. Put differently, we would like to analyze the dynamics behind a group of AI agents solving collectively a problem requiring ``some minimum amount of intelligence\footnote{ As there is no consensus on a formal definition of intelligence, we use the term in an abstract way to represent any cognitive or problem solving capability}''. The results from such a study might be very useful for developing collective intelligence tests, when combined methods for measuring individual intelligence.

Questions which are interesting in this context are for instance:
\begin{itemize}
\item Can the intelligence of a group be significantly higher than the intelligence of the (most intelligent) agents in the group (with positive influence due to collaboration)?
\item Can it be smaller (with negative influence through disturbance)?
\item How should the agents aggregate their skills? Which joint decision making system is the most suitable? Is it beneficial to provide the agents with a measure of their relative intelligence?
\item If several tasks must be performed simultaneously, how can the group allocate its resources? 
\end{itemize}
	
The approach will be mainly experimental as based on computer simulations, using models of collective behavior based on swarm intelligence, task allocation and problem specialization.

\section{Overview}

We will start defining the concept of intelligence, and present the main approaches and related formalisms for measuring it (chapter \ref{ch:intelligence}). 

Then we will present the concept of collective intelligence (chapter \ref{ch:collIntell}). We will mostly focus on collective intelligence for machines. However,  related concepts for humans such as the wisdom of the crowds and social intelligence will also be mentioned.

In chapter \ref{ch:approach} we explain the approach used to simulate a collective intelligence test. We will hence explain how we model the test problems and  how we introduce social aspects into the test.
Thereafter, we perform some experiments (chapter \ref{ch:experiments}). We have several test setups, each of which will be explained first, and then analyzed and interpreted.

In chapter \ref{ch:sumResults} we analyze the results from a more integrated perspective.

In chapter \ref{ch:futureWork} we explain some additional setups which might be interesting for future work.

Finally we conclude by looking at what objectives have been met and provide the take-away message from this report.

\chapter{Intelligence and intelligence tests}
\label{ch:intelligence}

In this chapter we will briefly define the concept of intelligence; both from the point of view of psychology and computer science. 

\section{Psychometrics, IQ tests and comparative cognition}

The concept of intelligence is fairly intuitive for all of us. It describes the ability of subjects -- humans, but also animals or machines -- to perform cognitive tasks. However, what exactly intelligence is, and which cognitive tasks precisely reflect intelligence, is less clear.

The most commonly accepted measure of intelligence is the so-called \textit{Intelligence Quotient} (IQ). This is in fact the normalized test score achieved in an IQ test. Hence, following \citet{Boring1923}, ``intelligence is the ability measured by the IQ test''. 

IQ tests are designed by psychologists; more precisely those working in the field of \textit{psychometrics}. Typically, the IQ test measures abstract reasoning capabilities. The problems of the test are mostly related to abilities such as verbal comprehension, word fluency, number facility, spatial visualization, associative memory, perceptual speed, reasoning, and induction. In 1904, the psychologist \citet{Spear04} discovered a positive correlation across the performances in these different tasks. He called the common factor in intelligence the general factor $g$. Consequently, the intelligence related to a specific type of task was denoted $s$. He argued hence that intelligence test should reflect the $g$-factor as it reflects the general ability of an individual to perform any cognitive task.

IQ tests are indeed fairly successful in discriminating humans according to who will be more or less successful in performing cognitive task encountered in real life. However, a problem of the tests is that they are too anthropocentric. It is thus badly suited for evaluating animals or machines. Moreover, some test problems of the IQ tests are those which are frequently faced by (adult) humans in real live and at which they are thus good at. As a result of this, it cannot be said that IQ tests reflect what might be understood as ``universal intelligence'', i.e., a valid concept for any kind of individual. A further discussion of what a universal test is and why IQ tests are not universal can be found in  \cite{IQnotformachines} and \cite{upsycho}.

The field of \textit{comparative cognition} extends intelligence beyond humans to animals. An important challenge of this discipline is that no human language or gestures can be used to provide the instructions for the test. To overcome this, rewards -- mostly in the form of food -- are used to incentivize the animals to achieve the highest score possible, hence to reveal its true ``intelligence''. These advancements have already provided us with interesting insights about the cognitive abilities of animals. (see for instance \cite{Shet09})

\section{The Turing test}

Also, with respect to the evaluation of machine intelligence, some advancement has been made. The \textit{Turing Test} expresses how far a computer is able to resemble a human. It is named after its famous inventor who was very concerned with machine intelligence already in 1950. \citet{Turing50} asked for instance whether computers would one day be able to ``think''. He was convinced that one day they would be able to do so. As the definition of ``thinking'' is again fairly abstract, he imagined the following test which he initially called it the \textit{imitation game}. In this test a human judge engages a written conversation with a human and a machine without knowing who is who.  If  in the future a machine would be build such that the judge cannot clearly distinguish the machine from the human, Turing argued that the machine could be said to``think'' and that it had attained the intelligence of humans. 

Over time, other versions of the Turing test have been invented so as to compare machines to humans. Some milestones of artificial intelligence could clearly be set this way. In 1997, a chess machine named \textit{deep blue} beat the Russian chess master Garry Kasparov \cite{Newb97}. In 2011, an IBM project called \textit{Watson} beat humans in the game Jeopardy! \cite{FLBGM12}.

Several tasks which initially allowed distinguishing machines from humans have over time lost their discriminative power. Currently, this can still be done using the \textit{CAPTCHA}\footnote{Completely Automated Public Turing test to tell Computers and Humans Apart} tests \cite{ABL02}, where  distorted letters and numbers have to be identified in a picture. This test is now omnipresent and is successfully used to avoid for instance the automated creation of email addresses by non intelligent machines. Yet, it is only a matter of time until this task can also be performed by machines.

It is hence becoming more and more difficult to distinguish humans from machines based on their performance in specific cognitive tasks. However, a good performance at the Turing test does not imply higher intelligence. This is related to the \textit{Chinese Room argument} brought forward by \citet{Searle80}. In his explanation, an operator disposing of a Chinese input-output table could maintain a seemingly intelligent conversation in Chinese, without actually being familiar with this language. And it has indeed been shown that a fairly easy algorithm could actually maintain a human-like conversation, without actually understanding the content of the conversation \cite{Salon03}. 

Moreover, the Turing Test is again too anthropocentric, as the reference object of intelligence is  human. The Turing Test can thus also not be used as universal intelligence test.

Computer scientists have hence made efforts to provide a universal definition of what intelligence actually is. Their current approach uses concepts of \textit{algorithmic information theory}, which will be explained next.

\section{Inductive inference }
\label{sec:indInf}

A recent approach has been to use \textit{inductive inference} for designing intelligence test, hence for to measuring and defining intelligence \cite{HM98,HernandezOrallo00a,LH07}. In inductive inference tests, the evaluee --  a human, an animal or a machine -- has to observe a non-random sequence of characters $\boldsymbol{x}=\{x_1,x_2,\hdots,x_t \}$. Typically, but not necessarily, those characters are assumed to be bits. The evaluee has then to learn the underlying pattern of this sequence and start predicting the next symbol $x_{t+1}$. When from a certain moment on all predictions are correct, we say that the evaluee has learned to predict the sequence.

The advantage of using inductive inference is that we dispose of a formalism to describe it mathematically. This formalism stems from \textit{algorithmic information theory} developed by  \citet{Kolmo65},  \citet{Chait66} and \citet{Solo64}. Let us explain the related concepts and how they might intervene in evaluating intelligence.

In order to evaluate intelligence, the evaluee must of course be tested on several sequences. As one might intuitively understand, there are sequences which are more difficult to predict than others. For instance the sequence $\{111\hdots\}$ is fairly easy to predict, while $\{010011000111\hdots\}$ is already more difficult. A mathematical formalization of a sequence's ``difficulty'' is its \textit{Kolomorow Complexity}. The Kolomorow Complexity of a sequence $\boldsymbol x$ -- actually the amount of information contained in it -- is given by the size of the smallest program $q$ on a Universal Turing Machine $U$ so that the latter generates this sequence on output \cite{LV93}:
\begin{equation}
K_U (\boldsymbol{x})=\min_q⁡{|q|:U(q)=\boldsymbol{x}}
\end{equation}

The definition of a \textit{Universal Turing Machine} (UTM) is a Turing Machine that can simulate any other Turing machine if previously fed with the appropriate program.

One can show that this definition is actually independent (to an extend) of the Turing Machine which is used. The difference of the complexity as measured on two distinct machines is a constant independent of $ \boldsymbol{x}$: $K_{U_1} (\boldsymbol{x})=K_{U_2} (\boldsymbol{x})+\mathcal{O}(1)$. This is because any Universal Turing Machine can be simulated by another with program of fixed length. This is known as the \textit{invariance theorem}. 

The Kolomorov complexity is often also referred to as the \textit{Minimum Description Length} (MDL) \cite{Wallace-Boulton68,Wallace2005,Rissanen83}, which is the shortest string, which taken as an algorithm produces $\boldsymbol{x}$. 

Intelligence as defined here is hence the ability of predicting a non-random sequence depending on its (Kolomorov) complexity. \citet{HM98} have designed a test based on inductive sequence prediction. They show that their scores are actually closely related to the IQ. The advantage of this approach over an IQ test is however  that we do not rely on some arbitrarily defined test score. Instead, there is now a well defined mathematical concept behind it.

The use of the Kolomorov complexity shows that the idea of inductive sequence prediction is actually strongly related to that of compression. The underlying task behind sequence prediction is to find the shortest description behind a sequence, which is nothing else than compressing it. We know that a totally random sequence cannot be predicted. In accordance to this, information theory tells us that it can also not be compressed. \citet{HM98} refer hence to ``intelligence as the ability of compression'', although they argue that this direct connection needs to be further refined and developed (and led beyond inductive inference \cite{HernandezOrallo00b}).

The use of compression (and its mathematical counterpart, i.e. Kolomorov complexity) solves another potential problem with the use of prediction as a measure of intelligence. Consider the sequence  $\boldsymbol{x}=\{2,4,6,8\}$. Typically one would predict the next number to appear as being 10, because the $k$th item of the sequence is given by $2k$. However, the polynomial $2k^4  - 20k^3+ 70k^2- 98k+48$ follows also the same initial pattern \cite{LH07}. According to this, the next number to predict would be 58. An intelligence test would however interpret 10 as the correct answer.  This is due to \textit{Occams's Razor principle}: ``If there are alternative explanations for a phenomenon, then -- all other things being equal -- we should select the simplest one''. The origin of this principle is rather philosophical. Yet, algorithmic information theory provides a mathematical justification to it. 

\citet{Solo64} defined the \textit{a priori probability} that on any input on a Universal Turing Machine $U$ appears the string $\boldsymbol{x}$. He considers for this the set of all programs which on output provide $\boldsymbol{x}$. The un-normalized prior of $\boldsymbol{x}$ is given by:
\begin{equation}
P_U (\boldsymbol{x})=\sum_{q:U(q)=\boldsymbol{x}}2^{-|p|} 
\end{equation}

This definition is again independent of the UTM which is considered. It is easy to see that this probability is dominated by the shortest description, hence that of length $K_U(\boldsymbol{x})$:
\begin{equation}
P_U (\boldsymbol{x})=\mathcal{O}\left(2^{-K_U (\boldsymbol{x})} \right)
\end{equation}

This is also known as \textit{universal distribution}.

Thus by selecting the shortest description, one selects actually the most likely one. This justifies Occams's Razor principle, which is consequently also called \textit{Minimum Description Length} (MDL) \textit{principle}.

An advantage is also that a measure of intelligence base on prediction/\nolinebreak com\-pression overcomes the Chinese Room argument\cite{DoweHajek97a}. As there exist an infinite number of sequences to predict, a look-up table cannot be used. Instead, prediction must actually be based on understanding the pattern underlying the sequence.

A disadvantage of using the Kolomorov complexity to evaluate intelligence is that it can typically not be computed in finite time (due to the Halting problem). However, there exist computable approximations of it (e.g., the Levin's $Kt$ \cite{Levin73}).

The Kolomorov complexity is a very powerful tool, which can be used beyond simply describing the complexity of a sequence. By analogy, it can also used as a measure for the complexity of a whole testing environment $\mu$. The complexity of this environment is simply the length of the shortest input to a UTM simulating the latter. This idea and its use for evaluating intelligence was pioneered by \citeauthor{dobrev2000} \cite{dobrev2000,dobrev2005formal}. It was then further elaborated in a more elegant way by \citet{legg2005universal} using Markov Decision Processes and reinforcement learning. 

A testing environment is typically described as a stochastic (Markov decision) process in which at each time step the evaluee takes an observation $o_t$ from an observation space $\mathcal{O}$ and receives a reward $r_t$ form a reward space $\mathcal{R}$. The evaluee will then chose an action $a_t$ from an action space $\mathcal{A} $. The probability of the couple $o_t r_t$  -- which is called the \textit{perception} -- depends on all previous actions, observations and rewards: $\mu(o_t r_t |o_1 r_1 a_1 o_2 r_2 a_2  ...o_{t-1} r_{t-1} a_{t-1})$. In the case of an inductive sequence prediction test, the observation is simply the last symbol $x_t$, the reward might be chosen at 1 and 0 depending on whether or not this symbol was correctly predicted. The action is then to select the next symbol. Using this formalism more complex environments can be described. \citet{HDEH11} discuss how an environment with several evaluees can be constructed. This is for instance useful for the design of adversarial prediction problems, where one evaluee has to predict a sequence generated by the other, as in \cite{Hibb08,hidh2012turing}.

\citet{LH07} use the here presented formalism from algorithmic information theory to design a universal measure of intelligence. Let $r_t^{i,\mu}$ be the reward of evaluee $i$ in environment $\mu$  at round $t$. The expected cumulative reward of this evaluee in this environment is the defined as:
\begin{equation}
\bar{\Phi}(\mu,i):=E\left(\sum_{t=1}^\infty r_t^{i,\mu} \right)
\end{equation}

There is a problem with the convergence of this series, which is however resolved here by supposing that the agent has a finite life and/or the environment can only provide a finite total reward. 

The proposed measure of \textit{universal intelligence} is actually the average of this expected cumulative reward in all environments $\mu$ from the environment space $\mathcal{E}$, weighted by the prior probability of this environment $P_U (\mu)=2^{-K_U (\mu)}$:
\begin{equation} 	\label{eq:univIndivIntell}
\bar{\Upsilon}(i):=\sum_{\mu\in \mathcal{E}}2^{-K_U (\mu)} \bar{\Phi}(\mu,i)
\end{equation}

This measure is called universal not because it uses the universal probability distribution, but rather because it could be applied to any type of evaluee: humans, animals and machines. It is still dependent on the considered UTM. From the invariance theorem we know however that another UTM would give the same result.

Even though the use of the universal distribution seems intuitive, it might be criticized here. The universal distribution gives a very high weight to the simplest problems. There are also other problems to turn the above measure into an intelligence test. Some of these issues are addressed in \cite{HD10,Hibbard2009}. One recurrent one is the use of a universal distribution to choose among environments instead of an actual measure of difficulty.
As put forward by \citet{HDEH11} and as we will discuss furthermore (see e.g., \ref{ssec:uniformProbWeight} and \ref{sec:singlePeak}), it might very well be that intelligence is not a monotonic phenomenon in the problem's complexity i.e., a very intelligent being might actually underperform less intelligent ones on very easy problems. In order to take this into account, \citet{HDEH11} suggest for instance to define a minimum complexity of the problems. Alternatively, intelligence might be defined as the maximum complexity of problems at which the performance differs significantly from random.

\chapter{Collective intelligence}
\label{ch:collIntell}

In the previous chapter we have considered intelligence from an individual point of view; both for humans, animals and machines. For humans however, it is very restrictive to consider their intelligence in an environment in which they are alone. Multiple aspects of what one might consider as ``intelligence'' relate to the interaction with other humans. Soon in the development of intelligence tests, this idea emerged under the name of \textit{social intelligence}.

Also for machines the same holds, yet to a lesser extent. In Artificial Intelligence the idea emerged that a group of less intelligent agents can actually outperform one very intelligent agent. For this to be true, the group must of course put its resources together. The agents must hence interact. The study and design of interacting artificial intelligence systems is called \textit{multi-agent systems} or \textit{collective intelligence}.

For the sake of completeness we will briefly discuss social intelligence. Thereafter we discuss collective intelligence. Both notions must however not be confused. Social intelligence reflects the individual ability of a human to interact with other human beings. Collective intelligence relates to interactions of artificial intelligence agents. Yet, it refers rather to the resulting intelligence of the group rather than the capacity of the individual agent to interact with others. Nonetheless, as we will see, the more these agents have ``social'' abilities, the higher will also be the resulting collective intelligence.

\section{A few words on social intelligence}

The idea of \textit{social intelligence} \cite{Kihl87} was first mentioned by  \citet{Thorn20}. He distinguished three facets of intelligence: the ability to understand and manage ideas (abstract intelligence), concrete objects (mechanical intelligence) and people (social intelligence). 

Many authors have since then discussed social aspects of intelligence. \citet{Ver33} defines it as ``the ability to get along with people in general, social techniques or ease in society, knowledge of social matters, susceptibility to stimuli from other members of a group, as well as insight into temporary moods or underlying personality traits of stranger''. 

Soon, the first social intelligence tests emerged. The first of such tests was the \textit{George Washington Social Intelligence Test} (GWSIT) \cite{Moss27}.  This test is composed of a number of subtests, which are combined to provide an aggregated index of social intelligence. Aspects which are measured are:
\begin{itemize}
\item	Judgment of social situations 
\item	Memory of names and faces
\item	Observation of human behavior
\item	Recognition of the mental states behind words
\item	Recognition of mental states from facial expression
\item	Social information
\item	Sense of humor
\end{itemize}

However, this test soon came under criticism as it correlated much with abstract intelligence tests \cite{Hunt28}. Hence some authors (e.g., \cite{Wech58}) argued that ``social intelligence is nothing else than general intelligence applied to the social domain''. Yet, we would like to have a test which measures abilities distinct from cognitive abilities.  Therefore many other tests of social intelligence have been proposed.

\citet{Sull65} for instance developed a test which seems to withstand such criticism \cite{SWF71}. Their test represents the ``social ability to judge people with respect to feelings, motives, thoughts, intentions, attitudes, or other psychological dispositions which might affect an individual's social behavior ''. They define six cognitive abilities:
\begin{description}
\item[Cognition of behavioral units:] the ability to identify the internal mental states of individuals
\item[Cognition of behavioral classes:] the ability to group together other people's mental states on the basis of similarity
\item[Cognition of behavioral relations:] the ability to interpret meaningful connections among behavioral acts
\item[Cognition of behavioral systems:] the ability to interpret sequences of social behavior
\item[Cognition of behavioral transformations:] the ability to respond flexibly in interpreting changes in social behavior 
\item[Cognition of behavioral implications:] the ability to predict what will happen in an interpersonal situation.
\end{description}

\section{The wisdom of the crowds}

Let us leave social intelligence aside now for a moment and consider collective intelligence. As mentioned in the introduction of this chapter, the idea behind the latter is that a group of less intelligent agents can be more intelligent than one very intelligent agent. The principle applies however equally to humans, which is usually referred as the \textit{wisdom of the crowds}. One of the first of having exploited this principle was Francis Galton \cite{Gal07}. In 1907, Galton visited a fair, at which there was a contest whose participants had to guess the weight of an ox. The closest guess would win a prize. Galton managed to get his hands on the people's votes after the contest. Most participants' vote differed a lot from the real weight. Some participants estimated far below and others far above the ox's weight. Out of the 800 participants, nobody guessed the correct value.  However, when Galton computed the median vote -- which he referred to as the \textit{vox populi}, the voice of the people --, he found out that it was actually in a 0.8\% range of the real weight. His conclusion was exactly what we refer to as the wisdom of the crowds. 

The wisdom of the crowds is present in very important applications of real life. The efficiency of markets is based exactly on this principle. On financial markets, participants make their bids for an asset. This is nothing else than providing their estimate its value. In most cases those bids will not reflect the correct market value. Yet, the errors made by the many bidders compensate each other and finally the market price will reflect the intrinsic value of the asset.

\section{Collective intelligence systems}

In artificial intelligence, \textit{Collective intelligence} \cite{WT99} commonly refers to multi-agent systems (i.e., a large distributed collection of computational processes) with no centralized communication or control. This means that there is no ``master agent'' which manages the other agents. Instead, each agent is autonomous and takes its  own decisions. Typically, each agent is very simple. The ``collective intelligence'' results from the cooperation and coordination of a large number of agents. In many cases, the agents are identical, but this is actually not necessary. 

Together, the group intends to maximize a common objective, the \textit{world utility function}. The difficulty when designing collective intelligence systems is to design the individual behavior(s) and objective(s) of the agents so that the world utility is maximized. 

How difficult the collective intelligence design task is can be illustrated with the tragedy of the commons which might arise when a group of agents acting individually intents to maximize a world utility function.

The \textit{tragedy of the commons} refers to a social dilemma that arises in the exploitation of \textit{common pool resources}, i.e. resources which are used by several individuals, as for instance nature, public defense, security, and sometimes even information goods. It was first discussed by Garret Hardin \cite{Har68}. He uses as example for such a resource medieval land tenure in Europe. This land was already at that time called \textit{the commons} as it was accessible to everybody. Every farmer could freely let its cattle grass on these grounds. Hardin showed that this led to over-grassing of the commons. This is because letting one more cow grass on the commons brings actually a benefit -- under the form of more milk -- to its owner, while is has a negative impact on other farmers as the common resource -- the grass on the commons -- becomes more scarce and diminishes their production of milk. In other words, exploiting the commons has actually negative externalities to society. The dilemma behind this is that most common pool resources will be overexploited, even though - or rather because - each individual acts rationally, i.e. it takes into account only its own costs and benefits. The negative externalities to all the other individuals are not considered. It would nevertheless be socially desirable to take them into account.

Such a phenomenon might of course also arise in multi-agent systems. So as to avoid the tragedy of the commons, the agents might hence not be too ``selfish'' and must focus on the world utility. Here one understands again how important cooperation and coordination among the agents are.

Before presenting different approaches of collective intelligence let us make an important remark about the form the agents take. As one can imagine, the agents might take the form of isolated physical entities. A typical example of this would be a group of robots interaction with each other. However, it is not necessary that the agents are physically isolated from each other. One talks also about collective intelligence when the agents are actually embedded in the same computational unit. This is for instance the case when collective intelligence is used for optimization algorithms. In this case the agents are represented by several instances of simple algorithmic objects. Most examples of such algorithms stem from the field of \textit{swarm intelligence} which we discuss below.

\section{Approaches to collective intelligence design}

There exist several approaches on how to design collective intelligence systems. We will briefly cite a few examples here.

\subsection{Reinforcement learning}

A first approach is to use \textit{Reinforcement Learning} (RL). In this approach, one would define the reward so that it is partially composed of an individual utility and of the world utility. Yet in many cases, RL is not well suited due to the big size of the action-policy space \cite{WT99}. We will hence not further discuss this approach.

\subsection{Market-based mechanisms}

As mentioned above, markets are a perfect example for an application of the wisdom of the crowds (``human collective intelligence''). One approach to design collective intelligence systems is to represent the problem to be solved by the group as a market. 

As an example, let us briefly explain here the approach taken by \citet{CBTD00}. In this paper, a group of paint-booths -- the agents -- have to paint trucks coming out of an assembly line. The goal -- i.e., world function -- is to minimize the total makespan of the trucks. The trucks have to be painted in colors depending on the customer orders. Each paint booth disposes of all colors. Yet, switching from one color to another requires an additional time to flush out the old paint and fill the booth with the appropriate color. Hence the number of color switches should be minimized. 

\citet{CBTD00} present a market-based approach, where each agent $i$ makes a bid to paint truck $j$. The truck will be assigned to the highest bidder. The bid of agent $i$  for truck $j$ is given by:
\begin{equation}
B_i (j)=\dfrac{P\cdot w_j  \left(1+C\cdot e(i,j)\right)}{\Delta T^L}
\end{equation}
where $w_j$  is the priority of truck $j$ and $\Delta T$ is the time it would take to paint truck $j$. The value of $e(i,j)$ is 1 if the last truck in the queue of booth $i$ matches the color truck $j$ needs to be painted, and 0 if taking on truck $j$ requires a paint flush. The parameters $P$,$C$, and $L$ are used to set the relative importance of the three components $w_j$,  $e(i,j)$ and $\Delta T$, respectively.

The advantage of the market-based approach is that it can revert to extensive knowledge from economics, such as the theory of general equilibrium and game theory, or more precisely auction theory. 

The disadvantage of such an approach is that mechanisms based on rational individualistic agents can frequently be defeated by other more sophisticated algorithms. This is because they are not based on cooperation between the agents and are hence prone to market failures such as the tragedy of the commons explained above. And indeed,  \citet{CBTD00} also present an ant-algorithm which defeats the market-based version. Ant-based algorithms belong to the family of biologically inspired swarm intelligence systems, which we will discuss next. The ant-algorithm used by \citet{CBTD00} is the same as the one we present in \ref{sssec:taskAllocModel}.

\subsection{Swarm intelligence}

\textit{Swarm intelligence} is ``any attempt to design algorithms or distributed problem- solving devices inspired by the collective behavior of social insect colonies and other animal societies'' \cite{BDT99}. Such social insects can for instance be ants, termites, bees, or even birds and fishes.

Social insects are suited for the design of collective intelligence systems as most of them are based on collective behavior  without centralized control. They are organized in large populations of self-organized, simple agents.

Most importantly however, one can be inspired from social insects as they use simple but effective models of communication. More precisely, social insects -- and hence also the agents of swarm intelligence systems -- communicate in two ways. First, there is \textit{direct communication} with nearby insects. This communication can be established via physical contact (antennation), visual contact, chemical contact, etc. 

Second, there is \textit{indirect communication} via the environment itself. More precisely, there is indirect communication when one insect modifies the environment and the other responds to the new environment at a later time. This is called \textit{stigmergy}. For ants and termites for instance this is done via pheromones. While walking, ants and termites deposit pheromones on the ground. Other ants will follow this pheromone trail with a high probability. 

Let us illustrate via a simple example how stigmergy can be useful for the group. Suppose the following experiment: close to an ant nest appears a new food source. There are two ways to get to this food source, one short and one long path. It is observed that in the beginning both ways are taken with equal probability. Yet, after a short moment, the colony will use the shortest path almost exclusively. The reason behind is that while walking to the food source and back to the nest again, the ants release pheromones on the ground, which other ants (and themselves) tend follow. Yet, the amounts of pheromones will rapidly become higher on the shortest path for two reasons. First -- the time reason --, as the shortest path can be crossed in a shorter amount of time therefore the ants having crossed the shortest path are available earlier to place their pheromones again. Second -- the distance reason --, given (initially) the same share of ants on both paths, the density of pheromones distributed will be lower on the longest path (as the density of ants on it is also lower). Soon as the share of ants become higher on the shortest path, the difference in the pheromone density will even be reinforced by the fact that the pheromones evaporate and must constantly be renewed.

By inspiring oneself from social insects one can design different kinds of collective intelligence algorithms. It is for instance possible to solve the shortest path problem similarly as explained before on a simplified network with only two possible paths. How exactly both ways of communication are translated in the algorithm is entirely at the discretion of the algorithm designer. 

In the example of the shortest path one would create a colony of artificial ants (the agents) which build iteratively random paths through the network from the start to the goal. The paths they build depend however on the (artificial) pheromone on the edges; the more pheromones on an edge the more likely an ant will take this edge. These pheromones are represented by a so-called \textit{stigmergic variable} $\tau_{ij}$, which represents the amount of pheromones on the edge between the nodes $i$ and $j$. After each round the pheromones are partially evaporated and the ants deposit some new pheromones. 

As mentioned, with artificial ants some improvements can be made with respect to real ants. For instance, instead of having the ants distributing the same amount of pheromones on the edges they have taken on their paths, this amount of pheromones might depend on the quality of the build path, as expressed by the inverse of the path's total distance (i.e., the world utility to be minimized). Also one can allow only the ants which have built the best paths to update the pheromones. The information in the pheromones $\tau_{ij}$ can also be complemented with heuristic information $\eta_{ij}$  such as the inverse of the distance between node $i$ and $j$, $\eta_{ij}=\tfrac{1}{d_{ij}}$ . Swarm intelligence algorithms can also be combined with local search algorithms.

However, for the shortest path problem, ant colony algorithms have always lower performance than algorithms such as Dijkstra. Yet, for NP-hard problems -- for which no exact algorithm exists -- ant-colony algorithms algorithms (and swarm intelligence algorithms in general) represent useful meta-heuristics. Examples of NP-hard problems for which ant-colony algorithms have been successfully developed are for instance the traveling salesman problem \cite{DG97}, routing problems \cite{BHS97}, scheduling problems \cite{YSLW05} or partitioning \cite{KLS97} problems, 

Not only for optimization algorithms, but also for collective intelligence systems with physically separated agents, social insects can be an inspiring source for designers. The corresponding field concerned with designing groups of collaborating robots is called \textit{swarm robotics}. We will not enter the details on how one can build such a swarm of robots. Instead, we will illustrate some real life application of such collective intelligence systems.

Several research projects have started to assess the use of swarm robots for search and rescue (SAR) tasks after disasters \cite{MGF05, Dav02, KTNMT99}. The advantage of using robots for SAR tasks is of course that the lives of human rescuers are not put into danger by sending them into dangerous environment. However, many simple robots seem more promising than one very sophisticated one. This is because a collective intelligence system has multiplied resources and hence not one single point of failure. If one robot fails -- which is likely in environments after a disaster -- the performance of the group will barely be affected. Also one can use different types of robots; each type specialized in on specific task. Flying robots for instance are more suited for searching large areas, while ground based robots are more suited for moving and transporting objects.

\section{Results from research on human collective intelligence}
\label{sec:humanCollect}

As mentioned in the introduction, the aim of this report will be to study the dynamics behind artificial collective intelligence. This is a novelty in the academic research. Yet, similar studies have already been conducted with groups of humans. 

The main conclusions of such research are basically identical \cite{Mill10, WM11, WCP10}. When a group of humans faces a cognitive task, its performance is only very weakly correlated with the average or even the maximum intelligence of its members. Instead, social aspects of the group members explained the group's results. A good communication and collaboration among the member is more important than the IQ of the members. 

A first important factor is the social sensitivity of the group members. In groups with high collective intelligence, the members had high social sensitivity for each other: they paid attention to each other and asked questions. Also in such groups, the members performed very well in social intelligence test where it came down to reading the others' emotions. 

Moreover, in groups with high collective intelligence, the turn-taking was distributed equally. Groups in which the conversation was dominated by only a few individuals are typically underperforming. 

Surprisingly the share of women in the group correlates positively with the collective intelligence. This is because women tend to be socially more sensitive. However, some gender diversity is  still advantageous.

Also, the way in which rewards (payments) are distributed in the group (even punishing those that underperform), as well as the use of have been sugested. One very interesting study is \citet{KBKV12}, who use a so-called \textit{crowdsource platform} -- a platform allowing employers to connect to several job seekers who will execute small tasks against a little reward -- to evaluate collective intelligence. They use this platform to solve IQ tests and show that even small groups can do better than 99\% of the population.

To put a long story short: Social intelligence plays an important role in collective intelligence. How well a group communicates and cooperates is important for its performance, perhaps more  than the individual intelligence of its members. Our collective intelligence test which we are attempting to design for machines must take this into account. Finally, there is an important issue in collective intelligence which is present in any social organization. Groups need to handle a diversity of situations. A way to cope with this variety is by member specialization, originated and reinforced by an appropriate task allocation.

\chapter{An approach based on abstracted intelligence, vote aggregation and task allocation}
\label{ch:approach}

As mentioned in the goal statement (\ref{sec:goalStmt}), the broad context of this master thesis is to develop an intelligence test for groups of AI agents. Such a quest is of course by far too ambitious, especially given the state of advancement in the field of research on tests of individual intelligence. Therefore the objective of this master thesis will only be to provide some insights on the dynamics of collective intelligence, which might contribute to the development of such a test in the future.

Two aspects are critical for the development of a coherent collective intelligence test:
\begin{enumerate}
\item As we have seen in \ref{sec:indInf}, the research field analyzing intelligence in a formal way is tending more and more in the direction of defining intelligence as being the capability of processing information; more precisely compression. The intelligence test for collectives should hence be based on similar concepts as well. We will however go a step further and make an abstraction of which problem precisely the agents are facing. 
\item As we have concluded in \ref{sec:humanCollect}, the collective intelligence of a group results from social aspects between its members. Ideally the test should reflect ``social'' aspects of collective intelligence. A group of very intelligent, but uncooperative agents can certainly be used to resolve some complex (information processing) tasks. Yet, in this case one cannot talk about ``collective'' intelligence. Collective intelligence always refers to some notion of collaboration. Our test should take this into account.
\end{enumerate}

We will discuss both issues in the following.

\section{Conceptualization of ``intelligence''}

As mentioned (\ref{sec:indInf}), we have some answers for the first issue, that is, about how one can conceptualize ``intelligence''. The notion of intelligence must be related to the capability of processing/compressing information. The Kolmogorov complexity provides us with tools to derive a mathematical concept of difficulty, which can eventually be used for measuring intelligence. Yet, estimating the Kolmogorov complexity of objects is in itself fairly complicated (due to the halting problem). Also we would like to analyze any ability and not only intelligence. 

Therefore, we will make an abstraction of the kind of test-problem we use to evaluate the intelligence/capabilities of the group.  We will henceforth not talk anymore about ``intelligence'', but rather about ``ability''. This way, what we are actually modeling could be any kind of test (also, for instance a physical skill). Thus, the here defined approach is assumed to be independent from any method of measuring intelligence and can easily be adapted to future developments in this area.

We will first explain how we make an abstraction of the ability we are measuring. Thereafter, we will explain how one can aggregate the abilities of a group of agents.

\subsection{Abstraction from information processing capabilities: Item response functions}

We will mathematically conceptualize abilities making reference to the field of \textit{item response theory} (IRT) \cite{LNB68}. In IRT, probabilities are assigned about how a person responds to an item; here a test (which is actually a set of items).
It is used in psychometrics, where it is useful for instance to provide a framework to analyze how well an assessment works so as to interpret the results and to refine it.

For this, an \textit{item response function} is defined. We will present briefly here the \textit{three parameter logistic model} \cite{LNB68}. In this model, the probability that a person succeeds on an item is given by:
\begin{equation}
P(\alpha)=c+\frac{1-c}{1+\exp\left[⁡-a(\alpha-b)\right]}
\end{equation}
where $\alpha$ is the person's \textit{ability parameter} and $a$, $b$ and $c$ are the \textit{item parameters}. They have the following interpretation:
\begin{description}
\item[$a$:] The \textit{discrimination} (scale, slope) represents the maximum slope of the response function (with respect to the person's ability $\alpha$). In other words it reflects whether or not less able persons have indeed a lower chance of succeeding similarly to a very able person in the test.
\item[$b$:] The \textit{difficulty} (item location), $p(b)=\tfrac{1+c}{2}$, represents the half-way point between the minimum ($c$) and and the maximum (1) probability of succeeding. It is also the point where the slope is maximized: $P' (b)=a\cdot\frac{1-c}{4}$
\item[$c$:] The \textit{pseudo-guessing}, (chance, asymptotic minimum), reflects the probability that  a person with infinitely low ability guesses the correct answer:  $P(-\infty)=c$ 
\end{description}

We will follow a similar approach, which we will adapt to our machine intelligence context. First of all, we will henceforth not talk anymore about persons, but ``agents''. We will define some function $P_{\lambda}(\alpha)$, which reflects the probability that an agent with some ability (intelligence) $\alpha$ finds the correct solution to a binary test problem with difficulty $\lambda$. A binary test problem has only two answer possibilities; e.g. true/false or 0/1. Such problems are for instance two class classification or binary sequence prediction. We also assure that both answers are equally likely a priori.

\begin{figure}[tb!]
\centering
\includegraphics[width=0.8\textwidth]{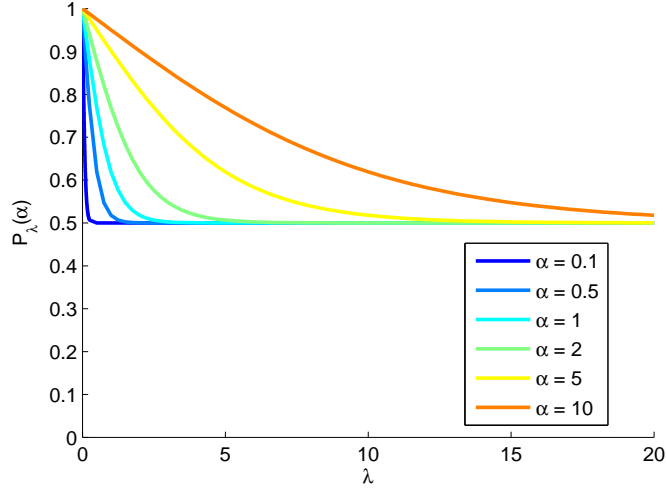}
\caption{A monotonically decreasing item response function $P_{\lambda}(\alpha)$ frem equation \eqref{eq:respFunc} for different values of the agent's ability $\alpha$}
\label{fig:respFunc}
\end{figure}

We will use here the function:
\begin{equation} \label{eq:respFunc}
P_{\lambda}(\alpha)=\frac{1}{2}+\frac{1}{1+\exp⁡\left[\tfrac{2\lambda}{\alpha}\right]}  ,
\end{equation}
which is shown in figure \ref{fig:respFunc} \footnote{ $P_{\lambda}(\alpha)=\frac{1}{2}\left(1+\exp⁡\left[\frac{-2\lambda}{\alpha}\right]\right)$  yields similar results (yet the slopes are steeper)}. We will henceforth refer to $P_{\lambda}(\alpha)$ as being the (expected) \textit{accuracy}.

The function has been designed so as to ensure that every agent has a 100\% chance to find the solution of a problem of no difficulty ($\lambda=0$). Thereafter, the probability of correctly solving the problem decreases monotonically with $\lambda$. Yet, it decreases slower the higher $\alpha$ is. As the difficulty of the problems increases, we suppose that the probability of solving the task converges to 50\%, i.e., the answer becomes random. In fact, letting the probability go down to 0\% (in the limit) would be less realistic. In this case, one would have to invert the provided answer in order to obtain an agent whose accuracy increases with the problem's difficulty. Yet, we discuss in our suggestion for future work what might happen when this probability goes below 50\% (See \ref{sec:asymPerfWorseRand} below).

The use of a function $P_{\lambda}(\alpha)$ is of course a generalization. The capability of an agent to solve a problem depends of course on the type of problem $\pi$ (e.g. sequence prediction, text recognition, speech recognition,$\hdots$ ). Typically one would rather define a probability function $P_\lambda(\pi,\alpha)$. Yet, we will make abstraction of this additional (and difficult to conceptualize) parameter. 

This simplification can be justified in two ways. First, we might suppose that the considered problem is the ``universal'' problem (or a set of all problems) allowing to measure ``intelligence'' as it might be defined in future research. Second, the function $P_{\lambda}(\alpha)$ might be considered as an average over different kinds of problems.

One should note that $\alpha$ cannot really be taken as an \textit{absolute} measure of abilities/intelligence. Yet, if $\alpha_i<\alpha_{i'}  \Rightarrow \forall\lambda: P_\lambda(\alpha_i)<P_\lambda(\alpha_{i'} ) $, then $\alpha$ can at least be used as a \textit{relative} measure of abilities/intelligence. The function $P_{\lambda}(\alpha)$ considered here in equation \eqref{eq:respFunc} has this property because it is monotonically non-decreasing for $\alpha$. Yet, as this is a fairly strong assumption we will also consider other types of functions $P_{\lambda}(\alpha)$ in our suggestions for future work (see \ref{sec:singlePeak}).

\subsection{Aggregation of a group's abilities: Voting systems}
\label{ssec:aggregateVoting}

As we are talking about collective intelligence here, we are considering not only one, but a group of $n>1$ agents of abilities $\boldsymbol{\alpha}=\{\alpha_1,\hdots,\alpha_n\}$. We will henceforth refer to the group, the set of agents, as $\mathcal{N}$. The individual agents are referred to by $i$. So as to measure the collective ability/intelligence resulting from the aggregation of  the groups, a joint decision making system is required. One of the most general configurations of such a system is for instance a \textit{voting scheme}.

The group will be faced with an evaluation problem of difficulty $\lambda$ (e.g., the prediction of a non-random series). Each agent will provide its answer $r_i$ to the problem (as determined by $P_{\lambda}(\alpha)$). The group might then use an \textit{absolute majority voting system} to determine its answer. The group will hence correctly solve the problem if more than 50\% of the agents have solved it correctly\footnote{ Supposing that we are in a case of a problem with a binary answer ($0/1$)}. In the case where exactly 50\% of the individuals solve the problem correctly -- and consequently the other 50\% provide the wrong answer -- the answer of the group is undetermined and will be drawn from a random coin flip.

The literature about multi-classifier systems \cite{Kunch04} provides us with some theoretical results about which performance might be expected from such a majority voting system. Suppose that all agents have the same ability $\alpha$, hence the same $P_{\lambda}(\alpha)$. Suppose also that the votes are independent of each other, which is somewhat restrictive here, as we cannot really talk about ``collective'' ability. In this case, the probability that the group solves correctly the problem (its accuracy) is given by:
\begin{equation}\label{eq:PmajGeneral}
P_{maj}=\sum_{k=\left \lfloor n/2\right \rfloor +1}^n \binom{n}{k} P_{\lambda}(\alpha)^k  \left(1-P_{\lambda}(\alpha) \right)^{n-k}
\end{equation}

Yet, when the agents have different abilities, hence different $P_{\lambda}(\alpha)$, we can only give an upper  and lower bound of the group's accuracy. For this, we first have to order the individuals by their individual accuracy -- which is in our case the same as sorting them by their abilities:
\begin{equation}
P_\lambda(\alpha_1)\leq P_\lambda(\alpha_2)\leq\hdots\leq P_\lambda(\alpha_n)). 
\end{equation}

Defining $k=\left \lfloor n/2\right \rfloor +1$, the bounds on the group's accuracy are given by \cite{Kunch04}:
\begin{align}
 \max ⁡P_{maj} &=\min⁡{\{1,\Sigma (k),\Sigma (k-1),\hdots,\Sigma (1)\}}   \\ \nonumber
 &\text{where } \quad  \Sigma (m)=\frac{1}{m} \sum_{i=1}^{n-k+m} P_\lambda(\alpha_i)  \\ \nonumber
 \min ⁡P_{maj} &=\max⁡{\{0,\xi(k),\xi(k-1),\hdots,\xi(1)\}}   \\ \nonumber
 &\text{where } \quad  \xi(m)=\frac{1}{m} \sum_{i=k-m+1}^n P_\lambda(\alpha_i)-\frac{n-k}{m} 
\end{align}

Whether the group is closer to the upper or to the lower bound depends on the complementarities of the agents. The upper bound can only be reached if the bad performance of some agents on some problems is systematically compensated by the good performance of other agents on these problems. On the opposite, if all agents perform similarly good or bad on the same problems, the lower bound is approached.

Majority voting systems are however not the only possible voting system. An interesting alternative are \textit{weighted voting systems} where the better agent will typically receive a higher weight. Again, the literature about multi-classifier systems \cite{Kunch04} provide us with some theoretical results about weighted voting systems. First, it can be expected, that the performance of the weighted version of the voting system performs better than the unweighted version. Within the different weighting schemes which might be used, weights proportional to $\log⁡\tfrac{P_\lambda(\alpha_i)}{1-P_\lambda(\alpha_i)}$ provide the best results.

We will hence define three voting systems for our experiments:
\begin{enumerate}
\item A majority weighting system where each agent has the same weight.
\item	A weighting system taking into account the abilities of the agents. We use hence a weight proportional to $\alpha$.
\item	The optimal weighting system using weights proportional to $\log⁡\frac{P_\lambda(\alpha_i)}{1-P_\lambda(\alpha_i)}$.
\end{enumerate}

Of course, so as to obtain a complete measure of the collective ability, we should observe the group's performance on a set of problems $\mathcal{M}$ with different difficulties $\lambda$ and aggregate the results using a weighted average: 
\begin{equation}\label{eq:weightPerfGroup}
\bar{\Phi}(\mathcal{M},n):=\sum_{j \in \mathcal {M}}w_{\mathcal{M}} (\lambda_j ) \bar{\Phi}(j,n)  ,
\end{equation}  
where $w_\mathcal{M} (\lambda_j )$ is a weighting function (thus $\sum_{j \in \mathcal {M}}w_\mathcal{M} (\lambda_j ) =1$) and \linebreak $\bar{\Phi}(j,n):= \bar{\Phi}(\alpha_1, \hdots, \alpha_n;\lambda_j)$ is the (empirically measured) average score/accuracy of the group on problem $j$ with difficulty $\lambda_j$. As mentioned in \ref{sec:indInf}, the current approach would be to express the difficulty by the Kolmogorov Complexity, $\lambda_j\propto K_U (j)$. Consequently, the weight would be represented by a universal distribution $w_{\mathcal{M} } (\lambda_j ):=2^{-K_U (j) }$. We have however already expressed our doubts about the use of a universal distribution in \ref{sec:indInf} and will hence not further discuss this for now and leave it over to \ref{ssec:uniformProbWeight}.

\section{Introducing social aspects}

We have discussed how to conceptualize intelligence as an abstract ability. We have done so using item response functions as abstraction, and voting systems as aggregation of abilities/intelligence. Yet, this aggregation is without any form of interaction, cooperation or specialization among the agents. The second issue is hence about taking social aspects into account. How to include social aspects in machine intelligence tests has until now been mostly unaddressed by the academic literature. The only exeption is a multi-agent extension of the individual intelligence tests performed in \citet{insa2012measuring}, where several cooperation and competition settings are studied. In our work, we are concerned about collective intelligence, and we will focus on how tasks are allocated.

\subsubsection{Testing on several problems}

As we mentioned in section \ref{sec:humanCollect}, one way in which a group of agents can improve performance over an individual is by a good allocation of tasks, mostly of this leads to specialization. Accordingly, instead of having to solve only one problem, the group faces several ($m\leq n$) problems of difficulties $\boldsymbol{\lambda}=\{\lambda_1,\lambda_2,\hdots\lambda_m \}$. We will henceforth refer to the set of problem as $\mathcal{M}$. The individual problems are referred to by $j$. 

Each of the $n$ agents can assign itself to only one of the problems at the same time. However, it can switch from one problem to another after each round. Among all the agents assigned to the same problem, the solution provided by the group will be determined via the aforementioned voting system. This way we require the group to coordinate/cooperate so as to distribute themselves among the various problems. 

Moreover, the problems have different levels of difficulty $\lambda$. Thus, ideally, more and/or the most able agents should work on the most difficult problems.

\subsubsection{Task allocation using the threshold model}
\label{sssec:taskAllocModel}

In order to make the group work, an algorithm must be proposed which allows the agents to allocate themselves to a problem without using a centralized decision maker (so that we can truly talk about ``collective'' intelligence).
One such algorithm is the \textit{dynamic task allocation algorithm} inspired from division of labor observed with social insects such as ants (\cite{BDT00, BSTD97, LLA04}). More precisely, it is inspired from an ant type called the \textit{Pheidole genus}. As observed by \citet{Wils84}, in such ant colonies two distinct types of ants are present. \textit{Minors} are occupied with day-to-day tasks such as breeding. \textit{Majors} take care of rather exceptional tasks such as defense. \citet{Wils84} observed however that if minors were retrieved from the colony, majors will consequently also perform the former's task. 

\citet{BTD96} showed that the division of labor/task allocation observed for the Pheidole genus, could easily be modeled trough a so-called \textit{threshold model}. 

In this model, a so-called \textit{stimulus} $S_j$ associated to the task $j$ is used. It represents the urge of performing a task, which must be translated into assigning more agents to a problem. For the real ant colony, the stimulus corresponds to some pheromones emitted by the ants. In our case $S_j$ might be a measure inversely related to the average performance over the last few rounds of the considered problem.

Also we define some \textit{response thresholds} $\theta_{ij}$ associated by each agent $i$ to each of the tasks $j$. This threshold reflects the matter of size above which the stimulus $S_j$  should be so that the agent $i$ will allocate itself to the corresponding task $j$. More precisely the probability that the later happens is given by the following \textit{sigmoid function}:
\begin{equation}
\mathrm{Prob}(i\mapsto j)=\frac{S_j^2}{S_j^2+\theta_{ij}^2 } ,
\end{equation}

\begin{figure}[tb!]
\centering
\includegraphics[width=0.8\textwidth]{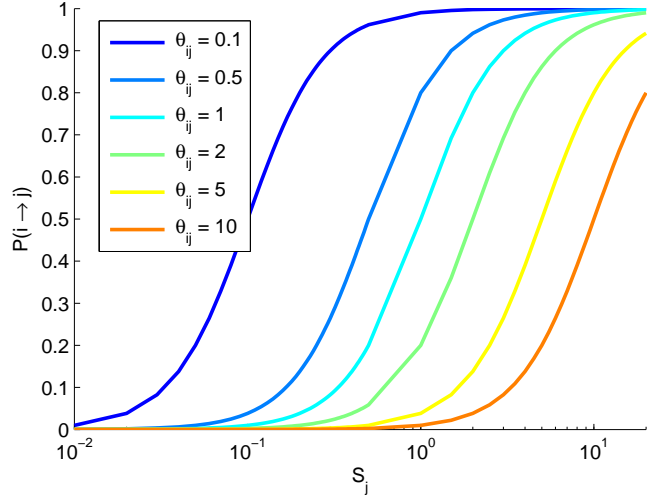}
\caption{The transition probability $\mathrm{Prob}(i\mapsto j)$ as function of the stimulus $S_j$ for various values of the response threshold $\theta$}
\label{fig:threshold}
\end{figure}

Such a sigmoid function can be seen as a continuous step (threshold) function. It is also used in other domains of computer science such as artificial neural networks. It is represented in figure  \ref{fig:threshold}. The squares are taken to give the function a faster transition from the lower to the higher probabilities.

In the real ant colony, thresholds are fixed and depend only on the type of ant. Minors have low threshold and will hence perform the breeding task most of the time. Majors have a high threshold and will only perform this task if the pheromones indicate a high urge to do so.


When being assigned to a problem, the agent will quit this problem with a probability $p$ per round and then select another one (or the same) to allocate itself using the previous rule. The expected number of rounds/steps spent on each problem before considering a switch is thus $\tfrac{1}{p}$, as the latter follows a geometric distribution. 

Both the thresholds and the stimuli vary over time. In the typical model, the stimulus is updated as follows: 
\begin{equation}
S_j\leftarrow S_j+\delta-\beta \frac{n_j}{n}  ,
\end{equation}
where $n_j$ is the number of active agents on task $j$, $n$ is the total number of agents, $\delta$ is the increase in stimulus intensity per time unit, and $\beta$ is a scale factor measuring the importance of the resources currently allocated to the problem.  The model tends therefore to distribute the workforce uniformly across the tasks. 

We however adapt this ``standard'' stimulus update rule to the situation in which it will be used. We can for instance make the stimuli dependent on the average performance of the last few time steps, in a way that the stimulus is higher when the group performed relatively badly on this problem.  More precisely, the average performance $\bar{\Psi}_j$ we use is represented by an exponentially moving average. 

The \textit{exponential moving average} \cite{WikiEMA} is obtained by attributing decreasing weights to the performances situated far in the past. The update rule for this average over N rounds is: 
\begin{equation} \label{eq:EMApsi}
\bar{\Psi}_j\leftarrow\omega_N \Psi_j+(1-\omega_N)\bar{\Psi}_j, 
\end{equation}
where  $0<\omega_N=\tfrac{2}{N+1}<1$. Of course, this average does not depend on a fixed number of rounds. It depends on by far more than $N$ rounds. Yet, more recent performance has a higher influence than that further in the past. More precisely, in the way the weight $\omega_N$  is defined, 86\% of the total weight is attributed to the $N$ most recent rounds.  

Also we include a factor $0<\zeta <1$ before the stimulus in order to prevent it to diverge. The rule becomes:
\begin{equation}\label{eq:stimulusUpdateFull}
S_j\leftarrow\zeta S_j+\delta-\beta \frac{n_j}{n}-\beta' \bar{\Psi}_j
\end{equation}
 
In addition, the tendency to distribute the agents uniformly among the problems must be put into doubt given the different complexities of the problems. We will hence test whether the term $\tfrac{n_j}{n}$ is useful in our case (see \ref{ssec:modelImprove}).

As mentioned, for the real ants the thresholds are fixed. In the standard model \cite{BDT00, BSTD97, LLA04}, the thresholds are updated in a way so as to avoid unnecessary switching from one task to another. This means that once an agent is attributed to task/problem $j$, its corresponding threshold will decrease, while that associated to all other tasks increases:
\begin{align} \label{eq:thresholdUpdateStandard}
\theta_{ij}&\leftarrow  \theta_{ij}-\xi \\
\theta_{ik}&\leftarrow  \theta_{ik}+\phi  \qquad  \forall k \neq j, \nonumber
\end{align}
where $\xi$ and $\phi$ are the \textit{learning} and \textit{forgetting coefficients}, respectively. The values of $\theta_{ij}$ are bounded from above and below: $\theta_{ij}\in[\theta_{min}  ,\theta_{max} ]$. 

Parametrizing this model might seem very difficult. Yet, we can inspire from \citet{BSTD97} to get a first idea of the parameters' matter of size. Finally, we propose the following parameter values:
\begin{align}
\theta_{min }&= 1,\quad & \theta_{max}&= 100,\quad & \beta&=\frac{m}{2}, \quad & \beta'&=4,  \quad & \delta &=4,  \\ 
p&=0.5,  			\quad &  \xi&=2,  			\quad & \phi&=0.5,  \quad & \zeta &=0.98,  \quad & \omega_N&=0.33 \nonumber
\end{align}

As can be observed, the coefficient $\beta$ corresponding to the share of allocated agents is dependent on the number of problems $m$. This is because the share of allocated agents is in itself dependent on this number. Typically, its matter of size is   $\tfrac{n_j}{n} \approx \tfrac{1}{m}$. We set therefore $\beta=\tfrac{m}{2}$ in order to make the whole term $\beta\tfrac{n_j}{n}$ independent of $m$.

The threshold allocation model is well suited for our need of a general example algorithm to perform the agent-problem allocation task. It is inspired from social insects and includes hence the communication methods typically used in the swarm intelligence design approach for collective intelligence systems. Some variables such as the stimuli and the model parameters are shared among the agents, which stands for some social interaction between them. Also, it is well described in the academic literature. As we will further highlight in \ref{ssec:dynEnvi}, the model is also able to allocate the agents appropriately in a changing (i.e., dynamic) environment. Moreover, the model can easily be adapted to our specific experimental needs.

\section{Factors of interest}

As we mentioned in the introduction, we are interested in evaluating collective behavior where some of the following features (or all of them) are considered:
\begin{description}
\item[Number of agents:] the number of agents is naturally one of the most important features to be considered.
\item[Agents abilities:] the degree of competence of the agents (and their diversity) is key.
\item[Number and variety of tasks:] we could have considered one task at a time which has to be solved by all the agents. While this will be one of the considered settings, we will explore the much richer problem of the group having to solve {\it several tasks at a time}. This implies allocation and, if tasks are different, specialization.
\item[Agent specialization:] if different tasks are used, should we have specialized agents?
\item[Collective decision making methods:] if there are more agents than tasks, then some agents will collaborate on the same task. Collective decision making policies will then be key here.
\end{description}
Overall, this is the first work which studies the problem of collective decision making and collective task allocation together, in the context of performance evaluation.

\chapter{Experiments}
\label{ch:experiments}

Given the approach defined in the previous chapter, we will perform some experiments so as to observe some interesting dynamics in collective intelligence as announced in the goal statement (\ref{sec:goalStmt}).

\section{Voting on one problem}
\label{sec:voteOneProb}
	
First of all, we will run the simulations using only one problem. This way we exclude the impact of the resource allocation algorithm and focus on the design of the voting system. What we would like to observe is how the group performs as compared to the average performance of the agents when facing the problem alone. A first thing to observe is the performance of a perfectly homogeneous group. As we will show in a theoretical way, we expect that the group can achieve any level of performance, given that the agents perform only slightly better than random, that there are a sufficiently high number of agents and that the agents' votes are independent. Yet, when the group is composed of agents with very different levels of intelligence we might observe other, more surprising results. We will for instance simulate a situation where one very intelligent agent interacts with many less intelligent agents (and vice versa).

\subsection{Homogeneous group of agents}
\label{ssec:singleProbHomogen}

To start with, we will imagine a very simple setup. There will be only one problem to be solved by various agents with the same ability $\alpha$. 

One can expect that the more agents work on the same problem, the better the accuracy of the group will be. More precisely, in a homogeneous group of agents, the stochastic process composed of the (independent) answers provided follows a binomial distribution: $r_1,r_2,\hdots r_n \sim B(n,P_{\lambda}(\alpha) )$

Hence the probability that among the $n$ provided answers, $k$ are correct is given by (see equation \eqref{eq:PmajGeneral})
\begin{equation}
\mathrm{Prob}( k)=\binom{n}{k} P_{\lambda}(\alpha)^k (1-P_{\lambda}(\alpha) )^{n-k}  
\end{equation}

The average share of individuals providing the correct answer will be $P_{\lambda}(\alpha)>50\%$ with a variance of  $\sigma ^2=\frac{P_{\lambda}(\alpha)(1-P_{\lambda}(\alpha))}{n}$. There exists always a number of agents $n$, so that the global accuracy is higher than any desired level. In fact, suppose that we would like to achieve an accuracy higher than $1-\gamma$. For this, the lower bound of the accuracy's one-sided confidence interval with confidence $1-\gamma$ must be above 50\%. In other words, we must be $1-\gamma$ percent sure that the majority provides the right answer.  Using the normal approximation\footnote{ The normal approximation of the binomial can be used, as we are in a process of independent and identically distributed random variables.} of the binomial distribution ($n\gtrsim 30$), this can be expressed as:
\begin{equation}
P_{\lambda}(\alpha)-z_\gamma \sqrt{\dfrac{(P_{\lambda}(\alpha)(1-P_{\lambda}(\alpha))}{n}}>50\%
\end{equation}
where $z_\gamma$ is the $\gamma$th quantile of the normal distribution. There is always an $n$ so that this is verified.

We will choose here a setup where the difficulty and the ability are in the same matter of size. This way we ensure that the agents are neither too good, nor too bad for the problem. More precisely we choose: $\lambda=\alpha=1$\footnote{The accuracy $P_{\lambda}(\alpha)$ of the agents is hence 61.9\% }.

\begin{figure}[tb!]
\centering
\includegraphics[width=0.8\textwidth]{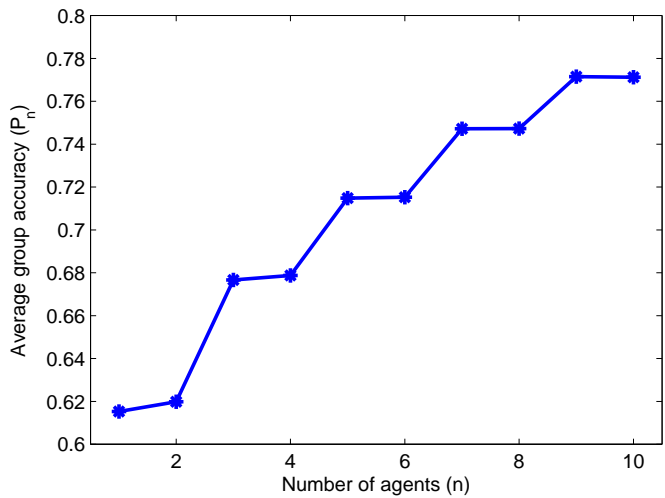}
\caption{Accuracy of a homogeneous group as a function of the number of agents on the problem ($\lambda=\alpha=1$)}
\label{fig:AsingleProblem}
\end{figure}

Figure \ref{fig:AsingleProblem} illustrates the explained phenomenon. We plot the average accuracy of the group $P_n$  over 30000 votes for a group composed from $n=1$ to $n=10$ agents. As expected the accuracy increases as the number of agents increases. 

Yet, one unexpected phenomenon can be observed. It seems that the accuracy of a group with an even number of agents is similar to that of the group with one agent less (i.e., the nearest and lower odd number). In order to get an intuition of what is happening, we will compare the accuracy of a group with one and two agents. Previously, we will simplify our notations: we define $p:=P_{\lambda}(\alpha)$ and $q:=1-P_{\lambda}(\alpha)$. The accuracy of one agent is of course equal to $P_1=p$. The accuracy of a group of two agents is composed of two terms: $P_2=p^2+\tfrac{1}{2} 2pq$. The first term reflects the probability that both agents are right; the second term the probability that only one of both agents is correct and that the coin flip necessary to determine the group's (random) answer is correct. This accuracy can be simplified to $P_2=p\cdot(p+q)=p$. Hence both accuracies are the same, $P_1=P_2$ \footnote{ Similarly, $P_3=p^3+3p^2q$ and $P_4=p^4+4p^3 q+\frac{1}{2} 6p^2 q^2=p^2\cdot(p+q)\cdot(p+3q)=P_3$}.

We demonstrate that this is true for any couple of an even number of agents and the nearest, lower odd number of agents. Actually what we state here is that:
\begin{equation}
\label{eq:PEvenOdd}
P_{2n}=P_{2n-1} \quad \forall n\in \mathbb{N}
\end{equation}
This is proven in appendix \ref{sec:ProofPEvenOdd}.

We have hence observed a first interesting dynamics in collective intelligence. In this simple setting, increasing the number of agents on a problem will increase the performance of the group. This is important to be considered when designing a collective intelligence test. Yet, in this particular case, one must make the distinction between an even and an odd number of agents working on a problem. It happens frequently -- as we will confirm as well for heterogeneous groups here below -- that going from an odd number of agents to the nearest superior even number does not increase the performance as much as going from an even number of agents to the nearest superior odd number. 

As we could understand from the developments above, this is due to the fact that a group with an even number of agents might find itself in a situation of an undetermined voting result, forcing it to use a coin flip to determine its answer. This has a negative impact on its performance and puts it at a disadvantage as compared to a group with an odd number of agents. This should also be considered when designing and interpreting the results of a collective machine intelligence test.

\subsection{The impact of adding low performing agents}
\label{ssec:lowPerfAgents}

We will now study a different setup with one problem. The aim here is to test the three voting systems we have defined: majority voting, voting weighted by $\alpha$ and using the optimal weight. In the previous point the voting systems all boiled down to majority (i.e., unweighted) voting as all agents were identical. We must hence put ourselves in a case of a heterogeneous group. 

\begin{figure}[tb!]
\centering
\begin{subfigure}[b]{0.8\textwidth}
        \centering
        \includegraphics[width=\textwidth]{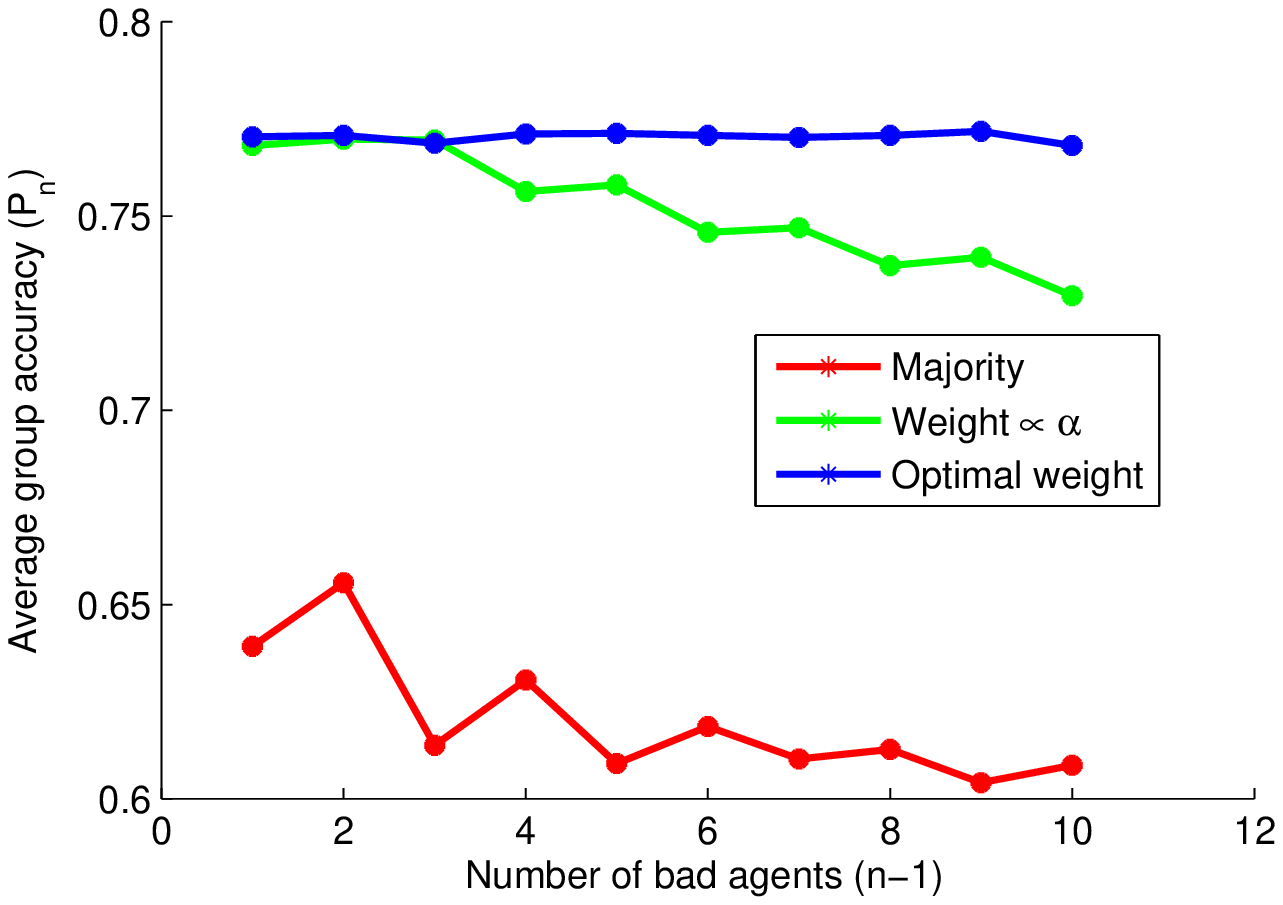}
        \caption{$\alpha_{bad}=5$}
\end{subfigure}
\begin{subfigure}[b]{0.8\textwidth}
        \centering
        \includegraphics[width=\textwidth]{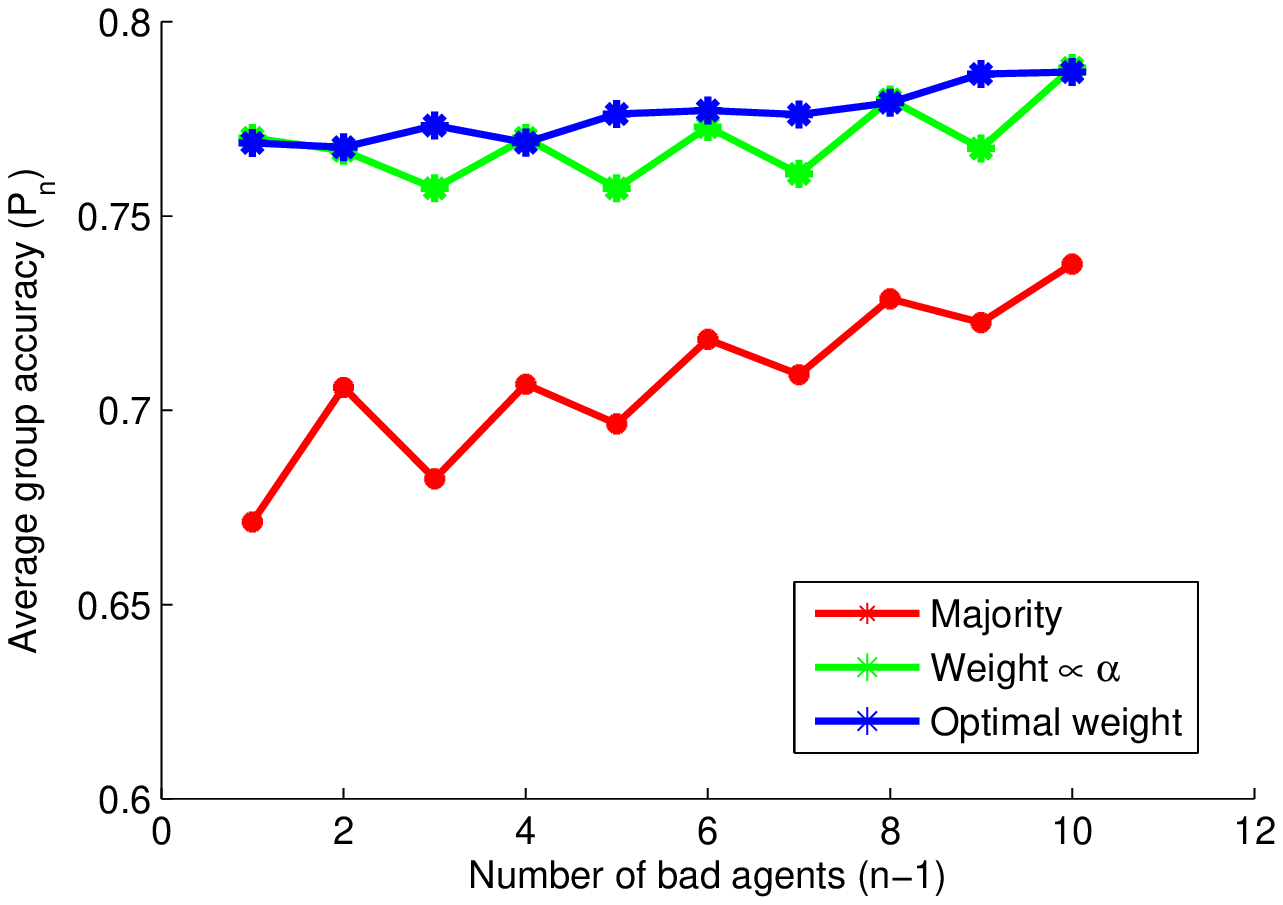}
        \caption{$\alpha_{bad}=8$}
\end{subfigure}
\caption{Effect of inserting bad agents for the three voting systems. There is always one good agent present with $\alpha_{good}=20$. ($\lambda=10$)}
\label{fig:AsingleProblemVoteBadAgents}
\end{figure}

We will study a group of very distinct agents; some have a very high ability\footnote{ As compared to the difficulty}, which provides them with an accuracy close to 100\%. Others have a very low one which makes their decision quasi-random. We will then observe how each of the joint decision making system is impacted when more and more agents with very low abilities are added. More precisely we define our unique problem as having a difficulty of $\lambda=10$. Our group always includes one ``good'' agent with ability $\alpha=20$\footnote{ The accuracy of such agents is 76.9\%}. Then we successively add ``bad'' agents with ability $\alpha=5$\footnote{ The accuracy of such agents is 51.8\%}.  The upper graph in  figure \ref{fig:AsingleProblemVoteBadAgents} shows the average accuracy of the group $P_n$  over 50'000 votes as a function of the number of bad agents for the three voting systems. The same experiment is repeated in the lower graph with $\alpha=8$ \footnote{ The accuracy of such agents is 57.6\%} for the bad agents. We will mainly discuss the upper graph.

A first thing we observe is that the performance of the majority voting system is significantly inferior to the performance of both weighted systems. It is also not surprising that the optimal weighting system has the best performance. Our results are hence coherent with what we know already form multi-classifier systems \cite{Kunch04}.

It can also be observed that the performance of the absolute weighting scheme decreases with the number of bad agents in the group. Yet, this decrease stagnates when adding more and more bad agents. This is because two distinct phenomena are underlying here. Adding underperforming agents is bad in the beginning as the weight of the high performing agent decreases and more weight is given to agents whose performance is close to random. At the same time, we know that a group of agents with an individual accuracy only slightly above 50\% can achieve any level of group accuracy given a high enough number of group members. Hence there is a compensating effect which increases the performance when we add bad agents. We know also that when the number of agents goes to infinity, the performance will increase up to 100\% for all three voting systems.

A similar, yet less pronounced effect of performance decrease is observed for the system weighted by $\alpha$. The system using the optimal weight seems completely unaffected by this phenomenon. It provides constant performance independently of the number of bad agents. This is due to the fact that the accuracy of the bad agents is so close to random that the weight associated to their votes is nearly zero.

In the lower graph of figure \ref{fig:AsingleProblemVoteBadAgents} we see that the performance decreases (or respectively stagnation) depends only on the very low performance of the bad agents. When this ability increases, the agents contribute positively in all three voting systems.

What might surprise us here is that we observe again a difference between an even and odd number of (bad) agents in the group. In the unweighted version we observe that including an even number of bad agents provides superior performance than an odd number. To see what happens we compare the situation when only one bad agent is present with that of a group including two bad agents. If only one bad agent is included, it happens frequently that the bad agent votes for the bad answer and the good agent for the good one. More precisely if one considers that the good agent is perfect (i.e., 100\% accuracy) and that the bad agent is fully random (i.e., 50\% accuracy), this situation occurs in exactly one case out of two. As the weights of both agents are identical, a coin flip must be used to determine the answer. When two bad agents are present, undetermined votes will never occur as there are in total an odd number of agents. Here it happens frequently that both agents have opposite votes and the good agent will provide the good answer. For one perfect and two random agents this will happen in 50\% of the cases. Only in 25\% of the cases the bad agents will both vote for the bad answer and outvote the correct answer of the good agent. The bad agents will hence ``cancel'' each other out and disturb the good agent less.

In the version weighted by $\alpha$ we observe that the performance of an odd number of bad agents is similar to the nearest lower even number of agents, $P_{2n-1}=P_{2n}$. Put differently, an even number of (all) agents is similar to the nearest lower odd number of agents. This is the phenomenon we observed and explained already in the previous point \ref{ssec:singleProbHomogen} for a homogenous group. We explained that this phenomenon is due to the fact that the performance of an even number of agents is reduced due to the necessity of using a coin flip in the case of an undetermined vote. Here we show that it also applies to a heterogeneous group. Yet, this is only possible as the ability -- hence the weight -- of the good agent $\alpha=20$ is a multiple of the bad agents' ability $\alpha=5$. Hence an undetermined vote is possible, however only for more than $\tfrac{20}{5}=4$ bad agents in the group. In the lower graph of figure \ref{fig:AsingleProblemVoteBadAgents} this is not the case anymore. Here we observe only the just explained effect that an odd number of bad agents disturb the good agent more.

\subsection{The impact of adding high performing agents}

\begin{figure}[tb!]
\centering
\includegraphics[width=0.8\textwidth]{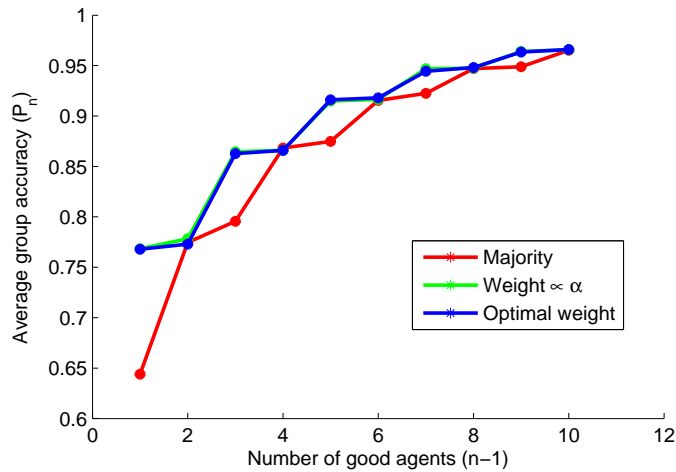}
\caption{Effect of inserting good agents ($\alpha_{good}=20$) for the three voting system. The group always includes one bad agent ($\alpha_{bad}=5$) ($\lambda=10$) }
\label{fig:AsingleProblemVoteGoodAgents}
\end{figure}

Figure \ref{fig:AsingleProblemVoteGoodAgents} shows the result for a similar experiment as the previous one. However here, we add good instead of bad agents. Trivially the performance increases for all three systems and converges to 100\% when we add more good agents. The performance of the weighted systems is superior or equal to the unweighted version. Both weighted versions have similar performances. Which exact weight is used is thus not of a great importance here. The only role of the weighting system is here to annihilate the influence that bad agents can have on the vote outcome.

What might surprise us is that the performance of a group including an even number of good agents is identical for all voting systems. This is because when the good agents disagree about the correct answer, the bad agent decides about the outcome of the group, whether the system is weighted or unweighted. 

Let us shortly also discuss what happens when the system includes an odd number of good agents. In the unweighted version it might happen that a coin flip is needed to determine the group's vote. Again, the one bad agent is more able to disturb the good ones. In the weighted version however a coin flip is never necessary.

\section{Simplified version of the allocation model with several problems}

In the previous section we analyzed the voting systems by working on only one problem. In this section we introduce for the first time a setup with several problems in order to verify the correct functioning of the allocation model and get familiar with it. However we will not immediately use the allocation model as defined in \ref{sssec:taskAllocModel}.  We will first use a simplified allocation model. Also we will modify our setup so as to impose an ``optimal'' allocation. This is to verify that our model is indeed capable of ``correctly'' allocating the agents.

We will first discuss the latter point and then explain how we simplify the model.

\subsection{Imposing an appropriate allocation: specialized agents}
\label{ssec:specializedAgents}

Until now we have not discussed how the agents should be ``optimally'' allocated. The question is hence how a group should allocate itself so as to maximize its performance. First of all, we need to define what we understand as performance measure. As mentioned above in equation \eqref{eq:weightPerfGroup}, the group's performance can be expressed as:
\begin{equation*}
\bar{\Phi}(\mathcal{M},n):=\sum_{j \in \mathcal {M}}w_{\mathcal{M}} (\lambda_j ) \bar{\Phi}(j,n)  
\end{equation*}

A difficulty with this expression consists in defining the ``right'' weight $w_\mathcal {M} (\lambda_j )$, as it is currently still an open question in individual intelligence tests (see \ref{sec:indInf}). There exist an infinite number of possibilities to define this function. 

Hence, we will make an abstraction of the aggregated measure of performance. Instead we define a more intuitive notion of the ``optimal'' allocation. What we will do is to associate one problem $\hat{j}_i$ to each agent $i$, such that the agent performs systematically better, say because the agent is a ``specialist'' in this problem. We will refer to agents which have been associated to a specific problem as ``specialized'' agents.

More precisely, what we will do is to increase the ability $\alpha$ for the agents' associated problems. Therefore, our abilities do not depend only on the agent $i$ anymore, but also on the considered problem $j$. Say that $\alpha_i$ is the ``typical'' ability as we have used it until now. Then the ability on the associated/specialized problem is just a multiple of this ability, i.e., $\alpha_{i\hat{j}_i }:=3\alpha_i$, while the ability on all other problems remains the same $\alpha_{ij}:=\alpha_i$.

This can be represented in the form of a matrix:
\begin{table}[h!]
\centering
\begin{tabular}{|cc|cccc|} \cline{3-6}
	\multicolumn{2}{c|}{ }  	& \multicolumn{4}{c|}{ Problem}  \\
	\multicolumn{2}{c|}{ }	&1	&2	&3	&4 \\ \hline
\multirow{6}{*}{\rotatebox{90}{ Agent}}	&1&	$3\alpha_i$&	$\alpha_i$&	$\alpha_i$&	$\alpha_i$	\\
		&2&	$3\alpha_i$&	$\alpha_i$&	$\alpha_i$&	$\alpha_i$	\\
		&3&	$\alpha_i$&	$3\alpha_i$&	$\alpha_i$&	$\alpha_i$	\\
		&4&	$\alpha_i$&	$\alpha_i$&	$3\alpha_i$&	$\alpha_i$	\\
		&5&	$\alpha_i$&	$\alpha_i$&	$3\alpha_i$&	$\alpha_i$	\\
		&6&	$\alpha_i$&	$\alpha_i$&	$\alpha_i$&	$3\alpha_i$ \\ \hline
\end{tabular}
\end{table}

One can understand that if the ability on an associated problem if sufficiently higher, the agent should be allocated to this problem\footnote{ A necessary condition for this is for instance that each problem has at least one associated agent. Otherwise this problem would be unserved and the group would hence achieve a performance of 0\% accuracy. Allocating at least one (un-specialized) agent would result in performance of at least 50\%} for most reasonable weighting functions. We have hence not proven that allocating each agent to its associated problem is optimal; more precisely, we cannot talk about optimality as we have not defined a weighting function. Yet, it is mostly ``appropriate'' to allocate an agent to the problems on which it performs systematically better. We will henceforth not talk anymore about the associated problem, but about the ``appropriate'' one. 

Of course, the table above can even be generalized to many other kinds of situations one would like to simulate, simply by modifying the pattern of $\alpha_{ij}$. One could for instance suppose that each agent has not only one but several appropriate problems.

\subsection{A simplified allocation model}

Our model as it has been defined now is very complex and includes many parameters. We need hence to simplify it. So as to eliminate a few parameters we will first work with a model, where the response thresholds $\theta_{ij}$  are static. In order to be mostly allocated to its appropriate problem, the response threshold of agent $i$ takes a low value on its appropriate problem and a high value on all others. 

We will also simplify the stimulus update rule. In order to eliminate at least one parameter, we will not consider the term $\beta \tfrac{n_j}{n}$ anymore. We have already put the usefulness of this term into doubt previously (\ref{sssec:taskAllocModel}). The update rule depends hence only on the average performance $\bar{\Psi}_j$ of the problem, which depends in turn on the appropriateness of the allocation. As a result, the simplified update rule is:
\begin{equation}
S_j\leftarrow\zeta S_j+\delta-\beta' \bar{\Psi}_j
\end{equation}

Concretely, this means that the stimulus to allocate more agents to a problem depends only on its current performance, and not on the share of agents allocated to it. More precisely, the lower the performance, the higher will be the stimulus to allocate more agents to the problem. 

\subsection{Simulation with specialized agents and the simplified allocation model}
\label{sec:simpleSpecializedSim}

\begin{figure}[tb!]
\centering
\includegraphics[width=0.8\textwidth]{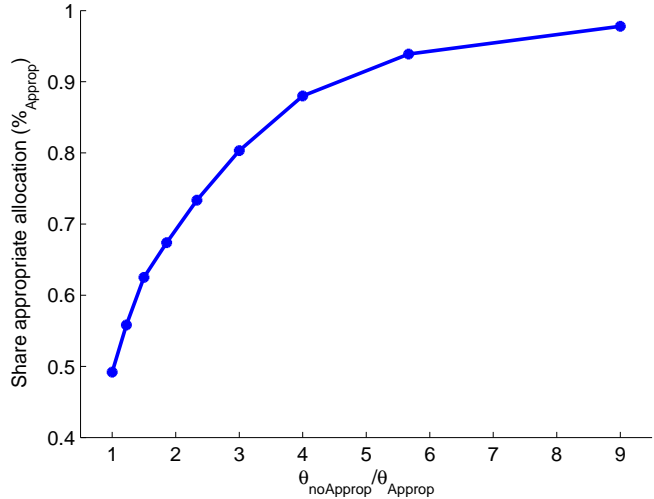}
\caption{Percentage of times the agent is appropriately allocated ($\%_{Approp}$) as a function of the ratio between the threshold on a non-appropriate problem and that of an appropriate problem $\tfrac{\theta_{noApprop}}{\theta_{Approp}}$ }
\label{fig:BassocSimpleThreshRatio}
\end{figure}

Figure \ref{fig:BassocSimpleThreshRatio} shows the percentage of the (discrete) time the agents are allocated to their appropriate problem -- henceforth noted by $\%_{Approp}$  -- as a function of the ratio $\tfrac{\theta_{noApprop}}{\theta_{Approp}}$, that is, the quotient of the threshold of a non-appropriate problems and that of the appropriate problem\footnote{ The exact values have been chosen so that: $\theta_{noApprop}+\theta_{Approp}=\theta_{max}$}. More precisely, the values shown here correspond to an average over 20 runs of a simulation with 1000 time steps. We will simulate two agents, each specialized in one distinct of the two problems.

This percentage of correct allocation is an interesting value to consider as it can be seen as a proxy for agent's performance independent of the problems' difficulties. As explained, the agents has a higher ability ($3\alpha$) on its appropriate problem, which is the reason why it is better to allocate it mostly to this specific problem. Hence, the average ability of the agent on its problems increases with this percentage: $\bar{\alpha}=3\alpha\%_{Approp}+\alpha(1-\%_{Approp} )$.    

Of course when $\tfrac{\theta_{noApprop}}{\theta_{Approp}}=1$, the agents are allocated randomly, that is 50\% of the time to their appropriate problem and 50\% to their non appropriate. Yet, as this ratio increases, the share of times the agents are appropriately allocated increases as well and becomes significantly different form a random allocation. For instance when $\theta_{noApprop}$ is three times bigger than $\theta_{Approp}$, the agents are on average correctly allocated in about 80\% of the time. 

Our allocation model works hence as expected and the agents are allocated according to the thresholds. 

\section{Standard version of the allocation model with specialized agents}

In this section, we will use the standard version of the threshold model so as to verify that the model is able to provide us with coherent values of the thresholds. We will still impose the appropriate allocation by specializing each agent to the problem. We will then adapt our model until we achieve the appropriate allocation. This way, we are sure that our parameter setting is correct. We will also discuss the introduction of the term $\beta \tfrac{n_j}{n}$ into the stimulus update rule.

\subsection{Allocation model testing and improvements}
\label{ssec:modelImprove}

\begin{figure}[tb!]
\centering
\includegraphics[width=0.8\textwidth]{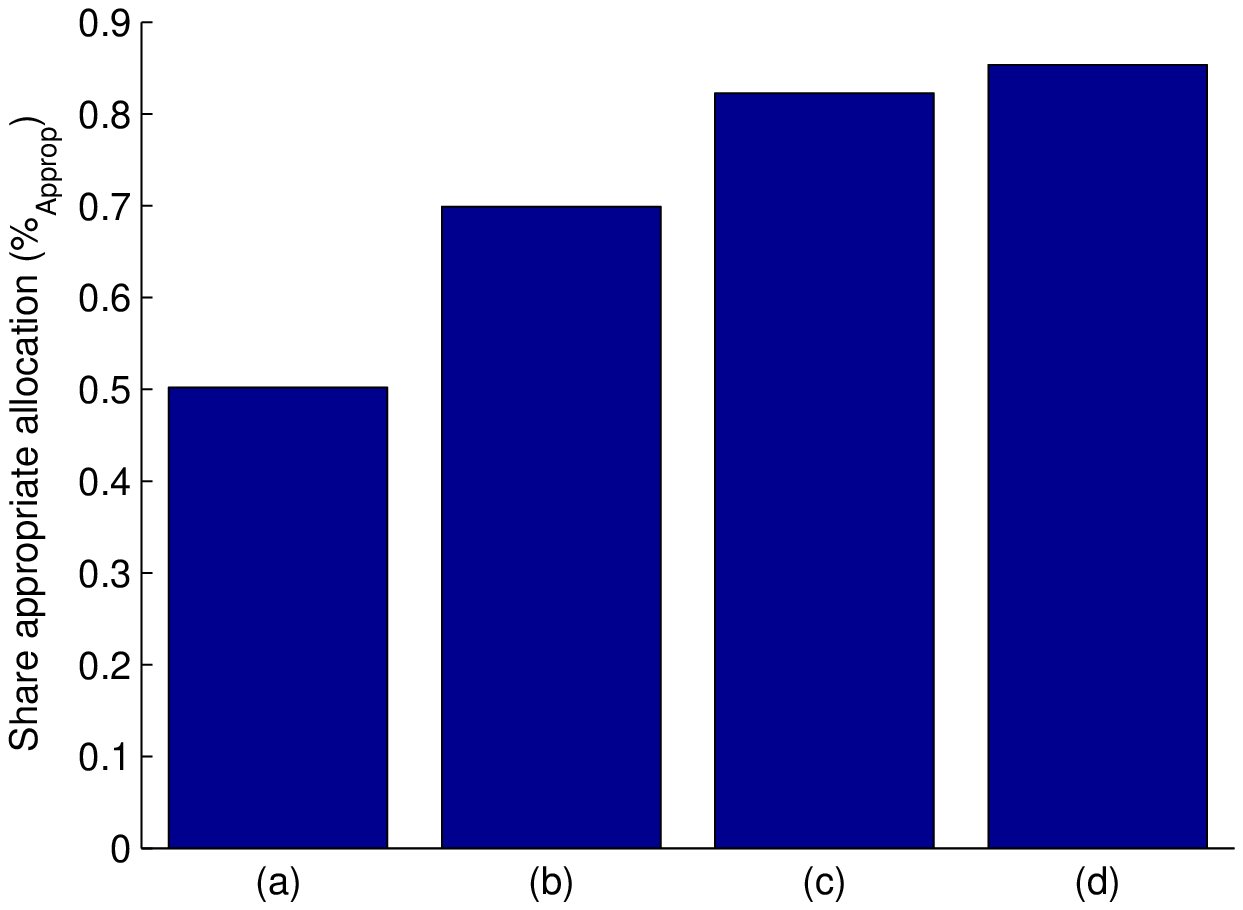}
\begin{tabular}{|p{5cm}|cccc|} 		\hline
Case							&(a) 	&(b)	&(c)	&(d)		\\	 \hline
Number of agents $n$		&2		&2		&4		&2		\\
Number of problems $m$	&2		&2		&2		&2		\\
Performance dependent threshold update rule	(equation \eqref{eq:thresholdUpdateImproved})	&No	&Yes	&Yes	&Yes		\\
Stimulus update rule dependent on  share of agents ($\tfrac{n_j}{n}$)   	&No	&No	&No	&Yes 		\\ \hline	
\end{tabular}
\caption{Time share of appropriate allocation $\%_{Approp}$ for four different settings}
\label{fig:CassocDynamic}
\end{figure}

The graph in figure \ref{fig:CassocDynamic} shows the share of correct allocations for the four the different settings we will discuss next. The values shown correspond to an average over 50 runs of a simulation with 1000 time steps. The simulations are made on 2 problems with difficulty $\lambda=10$. The specialized agents have an ability $\alpha=8$\footnote{ The accuracy of such agents is thus 57.6\% on a non appropriate problem and 80.3\% on their appropriate one}.

\subsubsection{(a) Standard model with 2 agents}
	
We use a setup with two specialized agents on two problems (the same as in the previous section). As we can observe, with the standard setting of the allocation model, the allocation does not differ from random. The agents are allocated 50\% of the time to one of the problems. As we have seen before, our allocation model works correctly once the threshold of the appropriate problem is lower than that of the non-appropriate problems. There must be a problem with our threshold update model in equation \eqref{eq:thresholdUpdateStandard}, which we repeat for convenience:
\begin{align*}
\theta_{ij}&\leftarrow  \theta_{ij}-\xi \\
\theta_{ik} &\leftarrow  \theta_{ik}+ \phi  \quad  \forall k \neq j,
\end{align*}
where $j$ is the problem agent $i$ has been allocated to.

Immediately we realize that nothing allows our thresholds to be systematically lower on the appropriate problem. The model has been designed to decrease the threshold on the problem the agent is currently allocated to, in order to avoid some unnecessary switching. Yet, this does not imply that the agent is also allocated to its appropriate problem.

\subsubsection{(b) Model depending on the individual performance with 2 agents}

\begin{figure}[tb!]
\centering 
\includegraphics[width=0.8\textwidth]{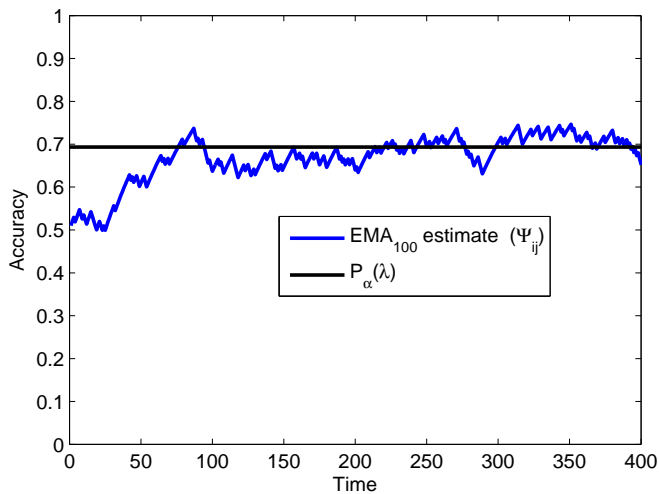}
\caption{Value of the accuracy estimate $\bar{\Psi}_{ij}$ as compared to $P_{\lambda_j}(\alpha_i)$ for an agent of $\alpha=7$ on a problem with $\lambda=5$}
\label{fig:EexpMaPerf}
\end{figure}

We must hence introduce a new feature to correct this. One might adapt the model so that it generates lower values for the appropriate problem. Yet, this would somewhat be ``too easy'' as the agents might for instance not know which is the problem they are specialized to i.e., the problem they are good at. We will consider here something we have not taken into account until now: the individual performance of the agent so far on each of the problems. We will estimate this as the exponentially moving average (see equation \eqref{eq:EMApsi})  of the accuracy over the last $N=100$ answers provided of agent $i$ on this problem $j$, noted $\bar{\Psi}_{ij}$. In figure \ref{fig:EexpMaPerf} one can observe that this empirical estimate of the accuracy converges indeed rapidly to the underlying accuracy $P_{\lambda_j}(\alpha_i)$.

We will use this measure so as to ensure that the agents have the tendency to be allocated to problems which they are good at and to be kept away from problems at which they are bad at. The model becomes:
\begin{align} \label{eq:thresholdUpdateImproved}
\theta_{ij}&\leftarrow\theta_{ij}-\xi\cdot\bar{\Psi}_{ij}^2  \\
\theta_{ik}&\leftarrow\theta_{ik}+ \frac{\phi}{\bar{\Psi}_{ik}^2}  \quad    \forall k \neq j \nonumber
\end{align}

The value of the threshold to which the agent is currently allocated to will thus be decreased faster, depending on how good the past performance on this problem was. At the same time problems to which the agents are currently not allocated, and on which they have performed relatively badly, will see their threshold increase faster. We take the square of our individual performance measure $\bar{\Psi}_{ij}^2$ so as to reinforce this effect, that is to have a better distinction between the higher values (close to 100\%) and the intermediate values (close to random i.e., 50\%) of $\bar{\Psi}_{ij}$.

The result is shown in the bar (b) of the figure \ref{fig:CassocDynamic}. Here we achieve a correct allocation in 70\% of the cases which distinguishes significantly from random. We obtain hence the desired allocation. Also intuitively, it seems natural to take the individual performance into account to allocate the agents.

\subsubsection{(c) Model depending on the individual performance with 4 agents}

Bar (c) in figure \ref{fig:CassocDynamic} shows the share of correct allocations when we use the previous model with four agents. One can observe that the fact of including more agents actually increases the likelihood that the agents are correctly allocated. This might be surprising in the first moment. 

Yet, there is reasonable explanation for this. If only two agents are working on the two problems, it happens frequently that both are allocated to the same problem. The other problem is hence un-served/abandoned which implies that the group's accuracy on it is at 0\%. In a perfectly random allocation this happens actually 50\% of the time. As a consequence of the resulting performance drop, the associated stimulus will increase rapidly and force an agent to be allocated to it. As the related threshold is still lower for the agent who has just left the problem, the latter will be reallocated to it. This phenomenon reduces hence the possibility that both agents exchange their problems.

However, when four agents work on the two problems, the likelihood that all agents will work on the same is actually reduced. In a random allocation, this happens in 12.5\% of the cases. This way a disequilibrium in the number of allocated agents might persist for a longer period of time. The agents can thus explore other problems. In the case that they perform relatively well on another problem they might allocate themselves permanently to it. Another agent -- actually the least performing -- will sooner or later leave this problem and in turn allocate itself to a problem at which it is better. This way a group of more agents is better in allocating itself in a more optimal way.

\subsubsection{(d) Model depending on the individual performance and share of allocated agents with 2 agents}

In equation \eqref{eq:stimulusUpdateFull}, we defined a stimulus update rule depending on the share of all agents which are allocated to a problem: $S_j \leftarrow S_j+\delta-\beta \tfrac{n_j}{n}-\beta' \bar{\Psi}_j $. Initially, we put the term $\beta \tfrac{n_j}{n}$ into doubt as it would result in a uniform distribution among the problems which might be undesirable. Yet, we see in bar (d) of the figure \ref{fig:CassocDynamic} that introducing this term -- yet with a smaller parameter than the one used for the problem's performance ($\beta=1<\beta'=4$)\footnote{ We have $m=2$, hence $\beta=\tfrac{m}{2}=1$. Typically it makes no sense to compare just the parameters. In fact, one has to compare the while terms $\beta \tfrac{n_j}{n}$  and $\beta' \bar{\Psi}_j$. Yet in our case $\tfrac{n_j}{n}$  and $\bar{\Psi}_j$ have the same matter of size - they are closely located at 50\%. Hence a comparison makes sense here.} - actually increases the likelihood that the two agents are allocated correctly. This might seem again very surprising.

The explanation of this phenomenon is however also related to that of the previous point. In fact, the inclusion of this term makes it easier for the agents to exchange their problems. As previously explained, when only two agents are working on two problems it happens frequently that the agents are allocated to the same problem. It happens then that the agent who joined this problem at last is reallocated to its initial problem as its threshold for the latter is still lower. Yet, with the inclusion of the term $\tfrac{n_j}{n}$ this issue is better resolved. In the case where both agents are on the same problem this term equals one on this problem, and zero on the unserved problem. This means that the stimulus for the unserved problem will decrease. At the same time, the stimulus for the served problem will decrease. This implies that both agents -- not only the one who joined last -- might go over to the unserved problem. Consequently, this mechanism makes it more likely that the agents exchange their problems so as to allocate themselves appropriately.

The inclusion of this term is hence not designed to ensure a uniform allocation of the agents. Rather its role is weaker through the lower associated parameter. It should be included in order to better handle situations in which some problem have no allocated agents and/or all agents are allocated to only one problem.

\subsection{Remark on dynamic environments}
\label{ssec:dynEnvi}

In the previous subsection we improved our allocation model and verified that it performs reasonably well i.e., the specialized agents are mostly allocated to their appropriate problem. Of course, our goal is not to achieve of the best performing allocation model. Yet, having a model which performs reasonably well is important to gain some insights from our experiments. 

It might be interesting to put forward one important aspect of our model: it is \textit{dynamic}. In the static version of our model we saw that we could get arbitrarily close to a 100\% appropriate allocation by fixing a high value of $\tfrac{\theta_{noApprop}}{\theta_{Approp}}$. In the dynamic version, this is not possible as the agents do not know which problem they are specialized on. They have to find it out by exploring the different problems. The fact of exploring regularly problems to which they should not be allocated reduces hence also the group's performance.

Yet, the dynamic version has one important advantage over its static equivalent: It is adapted to changing environments. Suppose for instance that after a few time steps, the difficulties of the problems change or that the agents are specialized on another problem. Such a dynamic environment could moreover be a very interesting feature of the testing environment for collective intelligence. It might be that two groups perform similarly in a static environment, but the one performs better in a dynamic environment. One might therefore conclude that this group is ``more intelligent/able'' than the other. A dynamic model is able to reallocate the agent adequately. 

\begin{figure}[tb!]
\centering 
\includegraphics[width=0.8\textwidth]{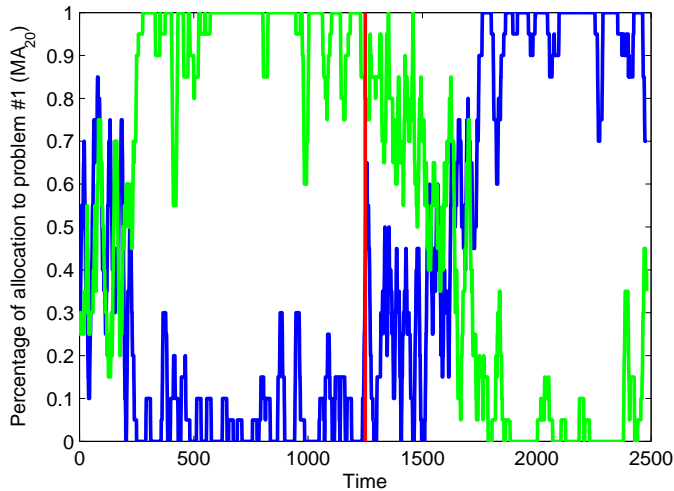}
\caption{Allocations of two specialized agents (blue and green) on two problems (0 and 1) over time. The red line indicates the exchange of the problem specializations}
\label{fig:CassocDynamicSwitchProb}
\end{figure}

In figure \ref{fig:CassocDynamicSwitchProb}, we perform the following experiment: We use again two specialized agents and two problems. We let the simulation run over 2500 time steps. After half the time -- indicated by the red line -- we exchange the association/specialization of the agents. On the $y$-axis we indicate for each agent whether it is allocated to problem 0 or problem 1. Yet, so as to smooth this curve the moving average over 20 periods is plotted. The blue agent is initially specialized on problem 0 and the green agent to problem 1.

We observe that there is first a phase of intense switching from one problem to the other, i.e., exploration. Then the agents become rapidly more and more allocated to their appropriate problem, i.e., exploitation. Yet, we observe still some attempts of problem switching. When the problem associations change, these attempts become more and more successful and last over a longer time. Finally, the agents completely allocate themselves to their new appropriate problem.

We have hence verified that our model has indeed this property of reacting correctly to a dynamic environment. This indicates that our parameters are chosen in a way that the thresholds and stimuli do not converge to extreme values which would make problem switching impossible.

\section{Use of the problems' difficulties in the allocation task}
\label{sec:diffPub}

Until now we have implicitly supposed that the agents do not know which of the problems are the easiest and which are the most difficult. Differently put, the agents did not use any information about the problems' difficulties $\lambda$ to allocate themselves. This will change now and the difficulties $\lambda$ become publicly known. We will modify the stimulus update rule in order to allow a higher share of allocation $\tfrac{n_j}{n}$ and to maintain a systematically higher stimulus on the most difficult problems.  
So as to simplify the setup -- yet without any loss of generality -- we will suppose that there exist three levels of difficulty:  $\lambda_l<\lambda_m< \lambda_h$. In the threshold model we might let the parameters of the stimuli update become dependent on the difficulty: 
\begin{equation}
S_j\leftarrow\zeta S_j+\delta_{\lambda_j }-\beta_{\lambda_j}  \frac{n_j}{n}-\beta' \bar{\Psi}_j 
\end{equation}

First, we will choose $\beta_{\lambda_l }>\beta_{\lambda_m }>\beta_{\lambda_h}$.  This implies that a higher share of agents can be allocated to the difficult problem without that the stimulus decreases significantly. The opposite holds for the easy problem to which we want to allocate the least share of our resources possible.  Also we choose $\delta_{\lambda_l}<\delta_{\lambda_m}<\delta_{\lambda_h}$  in order to force the stimulus to be systematically higher on the difficult problem, and vice versa.

The advantage of this approach is that we simulate a fairly general situation in a simplified way. Phenomena which will appear on three problems are typically the same as those on four or more. Also we can independently defined the three values of $\delta_{\lambda_j}$ and $\beta_{\lambda_j}$, without explicitly defining the function which links the difficulty $\lambda_j$ to these parameters; $\delta(\lambda_j)$ and $\beta(\lambda_j)$. However, for every triple of the of values for $\delta_{\lambda_j}$ and $\beta_{\lambda_j}$, there exist functions such that $\delta_{\lambda_j}=\delta(\lambda_j)$ and $\beta_{\lambda_j}=\beta(\lambda_j)$, respectively.

\subsection{Simulation}
\label{ssec:simDiffPub}

We perform the following experiment. Our group includes 12 homogeneous agents with ability $\alpha=5$. They have to be allocated to 3 problems with difficulties $\lambda_l=3; \lambda_m=5; \lambda_h=6$. Hence an individual agent would achieve an accuracy of 73.1\%, 61.9\% and 58.3\% respectively. 

\begin{figure}[tb!]
\centering 
\begin{subfigure}[b]{0.8\textwidth}
        \centering
        \includegraphics[width=\textwidth]{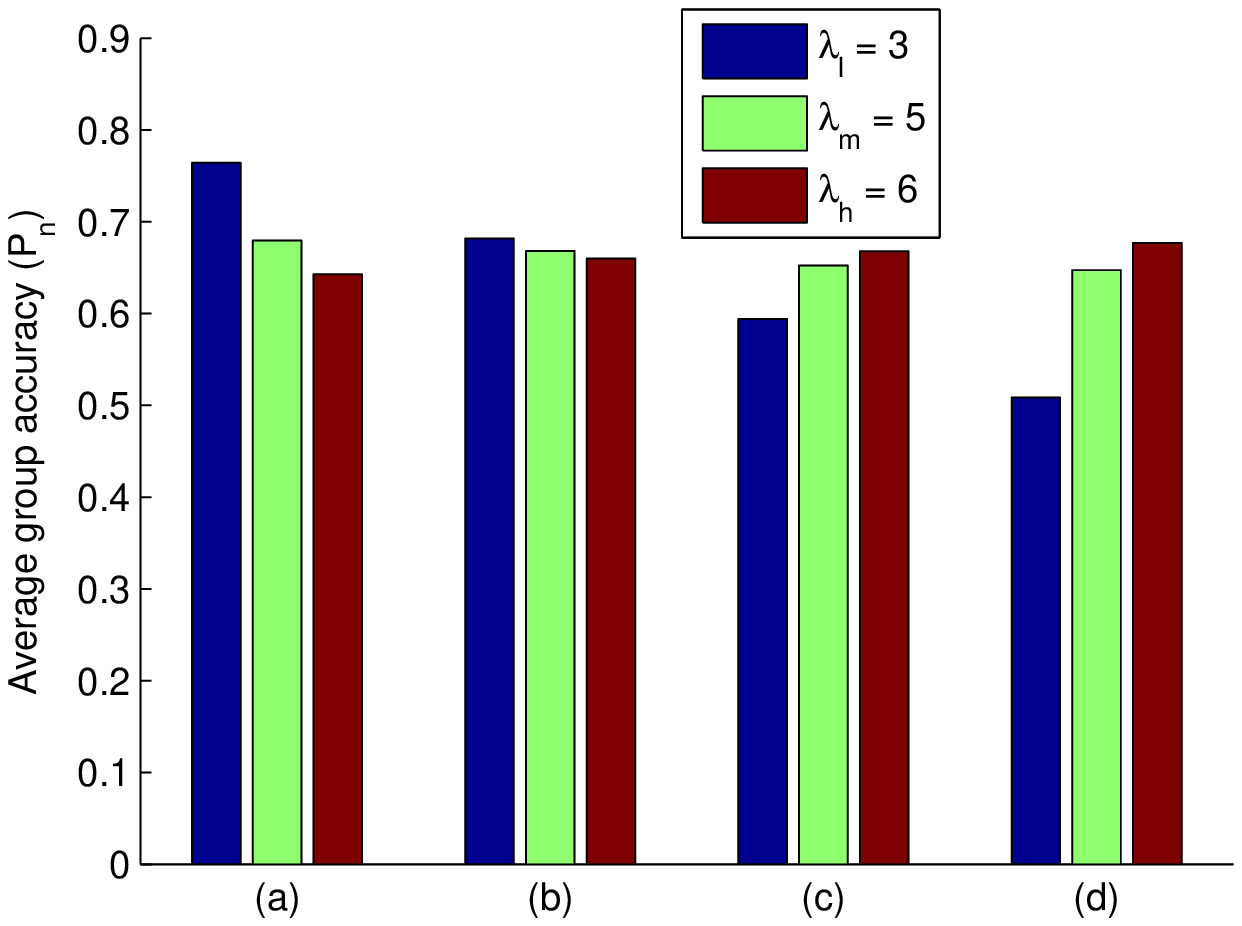}
\end{subfigure} 
\begin{subfigure}[b]{0.8\textwidth}
        \centering
        \includegraphics[width=\textwidth]{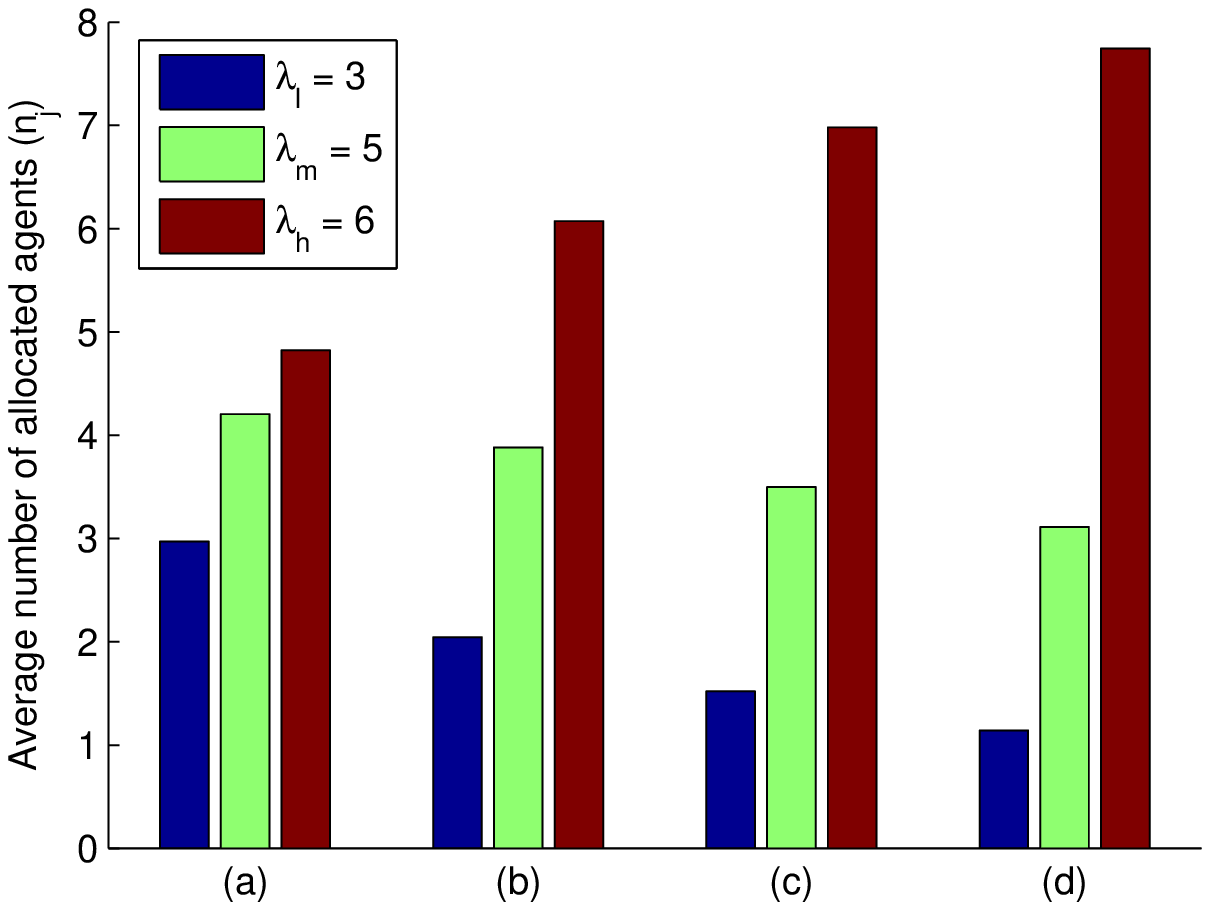}
\end{subfigure}
\begin{tabular}{|c|cccc|}\cline{2-5}
\multicolumn{1}{c|}{ }		&(a)		&(b)		&(c)		&(d)		\\ \hline
$f_{high}$						&1		&1.1	&1.2	&1.3	\\
$f_{low}$						&1		&0.9	&0.8	&0.7 	\\ \hline
\end{tabular}
\caption{Average group accuracy and average number of allocated agents for different parameter settings of a difficulty dependent stimulus update rule. The test includes three problems:  $\lambda_l=3$ (blue), $\lambda_m=5$ (green), $\lambda_h=6$ (red). There are $n=12$ agents with ability $\alpha=5$.}
\label{fig:DdiffPublic}
\end{figure}

The graphs in figure \ref{fig:DdiffPublic} shows the results of this experiment over 30 runs of a simulation with 1000 time steps. The upper graph shows the average group accuracy for the three problems.  In thelowere we show also the average number of agents allocated to each of the three problems.

We analyze four different cases of parameter settings. Case (a) serves as a reference where all parameters have the same values on the three problems. We use the typical parameter values $\delta=4$ and $\beta=\tfrac{m}{2}=1.5$. In the other cases multiply the parameters we want to increase ($\beta_{\lambda_l }$ and $\delta_{\lambda_h }$) by a factor $f_{high}>1$. Similarly we multiply the parameters we want to decrease ($\beta_{\lambda_h}$  and $\delta_{\lambda_l }$) by a factor $f_{low}<1$.

What one can observe is that we produce indeed the desired effect. We observe in the lower graph that as our parameters become more distinct on the problems, more agents are retrieved from the easy problem to be allocated to the most difficult one. As a consequence the performance on the former decreases considerably, while that on the latter increases -- due to its high difficulty -- only slowly. As we can see in (c) and (d), the parameters can even be chosen so as to achieve superior performance on the more difficult problems. Being able to achieve superior performance on the most difficult problems might be an important ability of the group, as a test might give higher importance to these problems.

Again, how to weight the importance of task difficulty is an open question. We will get back to this issue in subsection \ref{ssec:uniformProbWeight}.

\subsection{Should the group be provided with a measure of difficulty?}
\label{ssec:provideDiff}

We conclude thus  that providing the agents with a measure of difficulty can indeed help them to allocate their resources better among the problems. This is especially true if we consider that a test might value some problems more than others; whether the most difficult or rather the easiest. Typically, the group must be informed about how important each problem is. This would be done by providing a weighting function which we noted $w_\mathcal {M} (\lambda_j )$ in equation \eqref{eq:weightPerfGroup}.

Yet, we will argue that not providing the group with this information might actually be more reasonable. This is for two reasons. First, it is unclear which measure exactly should be provided to the group. More precisely, a measure of difficulty must always be relative to the agent. Second, an intelligent group does not need this information and can generate/explore all necessary information itself. We will discuss each argument in turn.

\subsubsection{Any agent independent difficulty measure is ambiguous}

Let us start with the first argument. Until now, our measure of difficulty was given by the parameter $\lambda$. As our response function is monotonically decreasing this parameter represents a valid \textit{relative} measure of difficulty. Mathematically this means that  $\forall\lambda_l< \lambda_h  ,\forall\alpha:P_\alpha (\lambda_l )<P_\alpha (\lambda_h )$. Yet, this is true for any monotonically increasing function applied to $\lambda$ (e.g. $\log⁡\lambda,\lambda^2,e^\lambda$). In the specific case of our simulation our measure of difficulty is well defined given the response function $P_{\lambda}(\alpha)$ from equation \eqref{eq:respFunc}. Yet, our response function $P_{\lambda}(\alpha)$ is just one arbitrary choice among an infinite number of other possible functions. Thus one cannot give a ``universal'' definition of difficulty which one should provide to the group.

Most importantly however, there exist many cases where $\lambda$ (and its monotonic transformations) are not even a relative measure of difficulty.  As mentioned, we have chosen $P_{\lambda}(\alpha)$  to be a monotonic function. Yet, there is reason to believe that there exist cases where this is not true anymore.  In \ref{sec:singlePeak}, we will present a model of single peaked response function. In this model each agent has a specific range of $\lambda$ at which he is good at. Yet, for problems with a significantly higher, and even lower values of $\lambda$, the agent performs significantly worse.  In this example $\lambda$ is definitely not an acceptable (relative) measure of the problems difficulty. 

Also using the formalism we have defined so far, we can easily generate situations where $\lambda$ ceases to be a valid measure of difficulty which would allow the group to define its allocation priorities. In \ref{ssec:specializedAgents} we considered specialized agents to whom we have assigned an appropriate problem at which they were good at. Also in such a case $\lambda$ ceased  being an acceptable (relative) measure of difficulty. This is because the agent's accuracy $P_{\lambda}(\alpha)$  does not depend on $\lambda$  exclusively anymore but also whether or not the agent is specialized on the considered problem.

Consider again the setup with 12 homogeneous agents we have just used in our simulations in \ref{ssec:simDiffPub}. Let us however introduce specialized agents here. More precisely, nine agents are specialized on the most difficult one ($\lambda_h=6$) and the remaining three are associated to the problem with intermediate difficulty ($\lambda_m=5$). In its typical parameter setting -- thus using the unmodified stimulus update rule in \eqref{eq:stimulusUpdateFull} --, the group performs best on the problems with the highest $\lambda$. More precisely the average group accuracies are: $P_n (\lambda_l )=83.9\%$; $P_n (\lambda_m )= 85.25\%$; $P_n (\lambda_h )= 87.0\%$. Hence $\lambda$ is not an adequate measure of difficulty for the group.

The problem with the difficulty measure $\lambda$ is that it is agent independent. It does hence not take into account phenomena such as non-monotonic response functions and agent specialization.

To sum up, we have shown, first, that there is no universal definition of measure of difficulty, and second, that there exist cases in which providing such a measure would be misleading for the group. A valid measure of difficulty must be specific to an agent.

\subsubsection{The environment itself provides sufficient information about the problems' ``difficulties''}

The second and supposedly most important argument why the tester should not provide a measure of difficulty is that the agents should actually find out which problems are the most difficult for them. Actually this could make the test even more discriminative. An intelligent group might rapidly find out which problems require more resources, while a less intelligent group would struggle perceiving the test environment correctly.

To show this, we will illustrate that the group can actually distribute its resources according to the problems' difficulties without receiving any indication about it. Let us define for this an objective. In case (b) of figure \ref{fig:DuniformPerformance}  we see that the group achieves nearly uniform performance on all problems. As this is an easily verifiable criterion, let us suppose that this is the group's objective. So as to attain this objective, the group must be aware of the problems' respective difficulty.  

Figure \ref{fig:DuniformPerformance} shows different attempts to uniformize the groups' performance. We state the average performance on all three problems over 30 runs with 1000 rounds. As a comparison, case (a) shows the performances with the typical parameter setting. Also case (b) relates to the performance which we have achieved by adapting the stimuli update parameter knowing the values of $\lambda$, as in case (b) of figure \ref{fig:DdiffPublic}.

\newcommand{\rott}[2]{
\begin{turn}{#1}
 \begin{minipage}[t][10pt][t]{4cm}
 	\begin{flushright}
 		#2
 	\end{flushright}
 \end{minipage} 
\end{turn}
} 
 
\begin{figure}[tb!]
\centering
\includegraphics[width=0.8\textwidth]{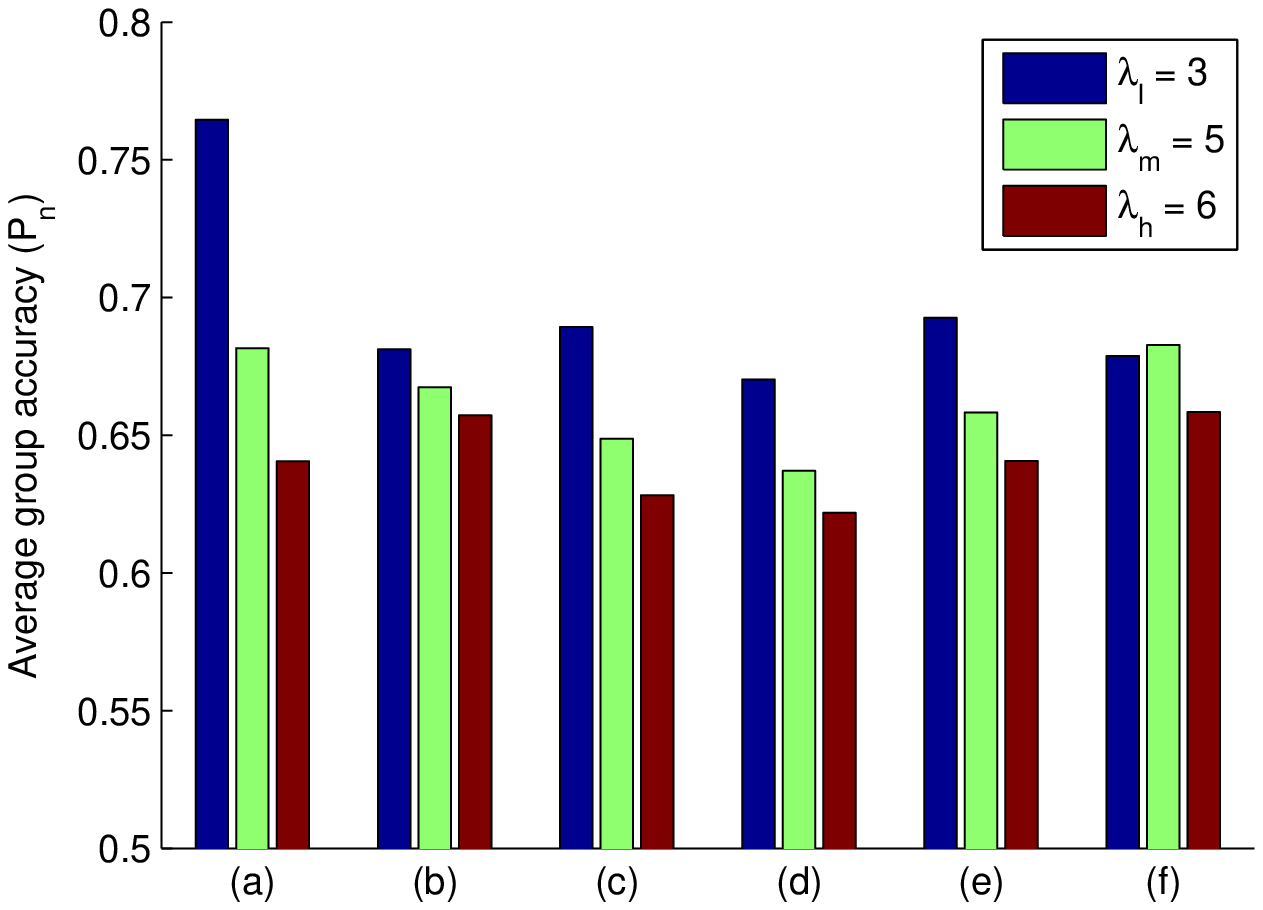}
\begin{tabular}[t]{|c|cccccc|} 		\cline{2-7}
\multicolumn{1}{c|}{ }			&(a)	&(b)	&(c)	&(d)	&(e)	&(f)		\\		\hline
$\delta$								&4		&4		&{\color{red}5}		&4		&2		&2.5		\\
$\beta$ ($\tfrac{n_j}{n}$)		&1		&1		&1		&1		&2		&2		\\
$\beta'$  ($\bar{\Psi}_j$) 	&4		&4		&{\color{red}7}		&8		&4		&4		\\
$f_{low}$							&1		&0.9	&1		&1		&1		&1		\\
$f_{high}$							&1		&1.1	&1		&1		&1		&1		\\ 		\hline
\raisebox{3.5cm}{$S_j\leftarrow$} &
\rott{80}{\small$\zeta S_j+\delta-\beta \frac{n_j}{n}-\beta' \bar{\Psi}_j \ $}	&
\rott{80}{\small$\zeta S_j+{\color{red}\delta_{\lambda_j}}-{\color{red}\beta_{\lambda_j } } \frac{n_j}{n}-\beta' \bar{\Psi}_j \ $}		&	
\rott{80}{\small$\zeta S_j+\delta-\beta \frac{n_j}{n}-\beta' \bar{\Psi}_j \ $}		&
\rott{80}{\small$\zeta S_j+\delta-\beta \frac{n_j}{n}-\beta' {\color{red}\bar{\Psi}_j^2} \ $}		&		
\rott{80}{\small$\zeta S_j+\frac{\delta}{\color{red}\bar{\Psi}_j}  -\beta \frac{n_j}{n} {\color{red}\bar{\Psi}_j}-\beta' \bar{\Psi}_j \ $}		&
\rott{80}{\small$\zeta S_j+\frac{\delta}{\color{red}\langle\bar{\Psi}_{ij} \rangle^p} -\beta\frac{n_j}{n} {\color{red}\langle\bar{\Psi}_{ij} \rangle^p} \ $ \\ $-\beta' \bar{\Psi}_j \ $  \hspace*{1cm} }		\\		\hline
\end{tabular}
\caption{Attempts to uniformize the performance across the problems}
\label{fig:DuniformPerformance}
\end{figure}

In our first attempt -- shown in (c) -- we give more emphasis to the problem's performance in our stimulus update rule. This is done simply by giving a higher value to the parameter $\beta'$ associated to the average performance over the last few steps on the problems $\bar{\Psi}_j$. In our second attempt -- shown in (d) -- we use the square of this performance estimate $\bar{\Psi}_j^2$. This way we give even more emphasis to the differences which might coexist between the problems performances. In (e) we make the values of $\delta$ (the constant term) and $\beta$  (the parameter associated to the share of agents allocated) dependent on the group's performance.  Similar to our approach taken in \ref{ssec:simDiffPub}, the value of $\delta$ should be high on difficult problems and $\beta$ should be low on difficult problems.  We use $\bar{\Psi}_j$ as a proxy for the problem's difficulty. Therefore we divide the former, and multiply the latter by $\bar{\Psi}_j$.

In all three cases we have empirically determined the parameter setting which makes the group's performance as uniform as possible on all three problems\footnote{ As expressed by the difference between the maximum and minimum performance.}.

Before discussing the results of our, one should note that the performance on all problems is inferior to that using the standard parameter setting. This is surprising as the performance decrease on one problem should be compensated by an increase on another one. The reason behind this is that by modifying our parameters in order to achieve a more uniform allocation, we have decreased the importance of the term reflecting the share of allocated agents ($\beta \tfrac{n_j}{n}$). As a consequence it happens more frequently that the problems are abandoned and achieve a performance of 0\%. We confirm hence again the interpretation we have given to this term in \ref{ssec:modelImprove}. 

We observe that our attempts to control the performance across the problems achieve only mediocre results. The reason behind this is that we use the wrong estimate for the problems difficulty: $\bar{\Psi}_j$. First, as $\bar{\Psi}_j$   reflects the very recent performance on the problem, it is a very fluctuating measure. Also this measure depends strongly on the number and abilities of the currently allocated agents and hence not only on the problem's difficulty. Suppose that at one moment the group achieves uniform performance on all problems over a few rounds. Hence, $\bar{\Psi}_j$ is the same for all problems and does not reflect the difficulty properly anymore.  In this case, the only term in the stimulus update rule which will differ across the problems is the share of allocated agents, $\tfrac{n_j}{n}$. As the performance is uniform, we know that this share must be higher on the most difficult problems. This will again force the group to allocate itself with a uniform number of agent across all problems. Agents will thus be taken away from the most difficult problem in favor of the easiest problems, which will in turn result in a non-uniform performance. Consequently, using $\bar{\Psi}_j$ as difficulty measure, a uniform performance cannot be a stable equilibrium of the allocation . 

Similarly to what we argue above with respect to $\lambda_j$ as a measure of difficulty, the biggest default of $\bar{\Psi}_j$  is that it does not reflect the difficulty as it is perceived by each of the agents. As we will further discuss below, it might be that each agent perceives the difficulty of a problem differently. It is for instance possible that one agent is good at a problem where another agent performs rather badly. However, for another problem the opposite might hold. Hence the difficulty is a very subjective measure and this must be taken into account. A measure reflecting the aggregate performance of several agents, such as $\bar{\Psi}_j$, is not acceptable.

What we need now is a better measure of the problems' difficulties, which can be easily estimated by the agents and which reflects their individual difficulty with the problem. In \ref{ssec:modelImprove} above we introduced $\bar{\Psi}_{ij}$, which is the exponentially moving average of  agent $i$'s accuracy on problem $j$. As mentioned, this measure converges rapidly to the underlying accuracy $P_{\lambda_j}(\alpha_i)$. Considering what we just said about a ``universal'' definition of problem difficulty, this measure is  acceptable as such, even though -- or rather because -- it depends on the considered agent. This measure\footnote{Note that $\bar{\Psi}_{ij}$ expresses rather the ``easiness'' of a problem, hence its inverse is the corresponding measure of difficulty} has a precise interpretation and takes specific characteristic of the agent (e.g., specialization in some types of problems) into account. 

Next we need to aggregate the agent-dependent measure of difficulty $\bar{\Psi}_{ij}$, into another measure which reflects the group's perception of the problem's difficulty, i.e., its priority to allocate its resources to it. This might be expressed as a generalized mean of $\bar{\Psi}_{ij}$ over all agents. The \textit{generalized mean} over all agents is given by: 
\begin{equation}
\langle\bar{\Psi}_{ij} \rangle^p=\left(\sum_{i=1}^n \bar{\Psi}_{ij}^p \right)^{\tfrac{1}{p}}
\end{equation}
where $p\in\mathbb{R}$. Using the generalized mean, many types of aggregations can be expressed and all of them make sense. 

For $p=-\infty$, the generalized mean is equals to the minimum of the terms $\langle\bar{\Psi}_{ij} \rangle^{-\infty}=\min_i⁡\bar{\Psi}_{ij}$ . We would hence use the accuracy of the worst agent on this problem as the group's measure of difficulty. For $p=+\infty$, the generalized mean is equals to the maximum of the terms  $\langle\bar{\Psi}_{ij} \rangle^{+\infty}=\max_i⁡\bar{\Psi}_{ij}$. We would hence use the accuracy of the best agent on this problem as the group's measure of difficulty. 

For all other values of $p$, the generalized mean lies between these two values $\langle\bar{\Psi}_{ij} \rangle^{-\infty}\leq\langle\bar{\Psi}_{ij} \rangle^p\leq\langle\bar{\Psi}_{ij} \rangle^{+\infty}$. It is increasing with $p$ : $\forall p_l< p_h:\langle\bar{\Psi}_{ij} \rangle^{p_l }\leq\langle\bar{\Psi}_{ij} \rangle^{p_h }$. The usual mean is of course obtained when $p=1$. For $p=0$, we obtain the geometric mean, which is more dependent on the lower values of  $\bar{\Psi}_{ij}$, hence the least performing agents. 

The question is now which value of $p$ the group should choose in order to properly aggregate the individual difficulties. As we will briefly show here, in some situations, $\langle\bar{\Psi}_{ij} \rangle^p$ should rather be close to the minimum individual accuracy, while in others it should be close to the maximum. Suppose for instance that all agents perform equally well on all problems except that there is one problem where only one of the agents performs well. Definitely the latter problem is the most difficult for the group and a measure close to the minimum accuracy reflects the group's ideal allocation priorities. Suppose now that all agents perform equally well on all problems except that there is one problem, where one of the agents performs extremely well. Definitely the latter problem is the least difficult for the group and a measure close to the maximum accuracy reflects the group's ideal allocation priorities.

Which value of $p$ should be chosen depends thus on the specific case (i.e., the form of $P_{\lambda}(\alpha)$ and which other factor (such as specializations) intervene in this function, the agents and the problems present). We will however not discuss how exactly $p$ should be chosen. This is left over to the ``intelligence'' of the group. Anyway in our case all values of $p$ provide us with the same mean as the group is homogeneous and thus $\bar{\Psi}_{ij}$ is the same for each agent.

In case (f) of figure \ref{fig:DuniformPerformance}, we use $\langle\bar{\Psi}_{ij} \rangle^p$ as a difficulty measure and we see that we achieve nearly uniform performance across the problems and this without the performance loss seen in the previous cases.  Choosing different values for the parameter of stimulus update rule, it has been verified that  the group could also achieve superior performance on the most difficult problems, similarly to what we observe in figure \ref{fig:DdiffPublic} when the values of $\lambda$ were public.

We have thus illustrated how the group can explore the environment in order to find out which problems are the most difficult and which deserve hence more resources. Of course, in general, this is not a trivial task for the group as it requires a high level of communication between the agents. Therefore it can -- and we argue that it should -- be used as a part of the intelligence test. The group should not be provided with a measure of difficulty, but find this out for itself.

\subsection{Uniform problem weighting}
\label{ssec:uniformProbWeight}

We just saw that by providing a measure of difficulty -- as expressed here by the response function parameter $\lambda$ --, but also by computing on its own a measure of difficulty $\langle\bar{\Psi}_{ij} \rangle^p$, the group could adapt its allocation model so as to take into account that some problems require more resources than others. More precisely, more difficult problems might be considered more important by the tester and might hence receive a higher weight $w_\mathcal {M} (\lambda_j)$ we used in our expression \eqref{eq:weightPerfGroup} of the group's performance: 
\begin{equation*}
\bar{\Phi}(\mathcal{M},n):=\sum_{j \in \mathcal {M}}w_{\mathcal{M}} (\lambda_j ) \bar{\Phi}(j,n) 
\end{equation*} 

Actually this weighting scheme $w_\mathcal {M} (\lambda_j )$ will again be important here and needs some further discussion. 

We just claimed that the tester should not provide the group with a measure of the problems difficulty $\lambda_j$. Yet, the importance of each problem, i.e., the weighting scheme $w_\mathcal {M} (\lambda_j )$ must be provided to the group. This is because in order to achieve the highest score  possible, the group must know how its results on the individual problems in $\mathcal{M}$  will be aggregated into one single performance measure $\bar{\Phi}(\mathcal{M},n)$. 

However, providing $w_\mathcal {M} (\lambda_j )$ to the group, while requiring that it has to find its own estimates of the $\lambda_j$ makes no sense. Actually, $w_\mathcal {M} (\lambda_j )$ contains a lot of information about the problem's difficulty. Let us take the example where the tester attributes more weight to the most difficult problems. In this case the function $w_\mathcal {M} (\lambda_j )$ is simply a (normalized) monotonically increasing function of $\lambda_j$. Yet, as we said before any monotonically increasing function applied to a (relative) measure of difficulty (suppose that it is valid in the specific case considered) is again a measure of difficulty. Hence, if we claim that the tester might not provide $\lambda_j$, then it can also not provide $w_\mathcal {M} (\lambda_j )$.

As a consequence the weighting of the problems importance must be uniform, as it is the only (non random) weighting scheme which provides no information about the problems difficulties. And there are actually some reasons which speak in favor of uniform weighting.

A first argument is related to what we said about dynamic versus static environments. In \ref{ssec:dynEnvi} we argued that dynamic environments are more interesting for testing, as two groups might perform similarly in a static, but differently in a dynamic environment. An interesting type of dynamic environments is of course those where the problems' difficulties change over time and where the agents have to reallocate themselves accordingly. If one would weight the problems according to their difficulty, one would have to re-weight them after the difficulties change. Consequently, one would have to communicate the weight change to the group. This is of course be a perfect indication to the group that the environment has changed. Based on this indication, a group could hence adapt itself to the new environment and perform well, even though it would have performed badly without this indication. Choosing a uniform weighting is therefore useful for disguise to the group any change in a dynamic environment.

Also we argued above that there exists -- at the moment -- no ``universal'' definition of the problems difficulty\footnote{For intelligence tests, a difficulty measure derived from the Kolmogorov complexity might become this definition}. Hence the weighting function is also undefined. If the exact weights cannot be defined, a uniform weighting is an easy way to avoid this difficulty.

Moreover, it seems reasonable to suppose that an ``intelligent'' group performs well at a variety of problems. More precisely, it should perform well on difficult problems, but also on easier ones. A less intelligent group will only perform well at the easy problems. A uniform weighting system reflects this requirement, while it is still able to discriminate groups of different abilities.

However, one might argue that non-uniform weights are still required for a test in order to discriminate properly. Uniform weighting is badly suited in a situation where a very intelligent agent performs badly on an easy problem and is thus outperformed by a less intelligent one. An explanation of this is for instance that the easy problem is actually ``too easy'' for the intelligent agent. In \ref{sec:singlePeak} below we present an example illustrating such a case. Suppose that our set of problems $\mathcal{M}$ contains one easy and one difficult problem. Hence, the intelligent and the less intelligent agent perform well on one of both problems. In this case, a test using a uniform weight might not discriminate both agents or even reveal the wrong agent as being the most intelligent. More weight must definitely be given to the most difficult problem. Yet, a higher importance can also be given to a specific type of problem by including more instances of it into the set of problems $\mathcal {M}$. By including more instances of the most discriminative problems into $\mathcal {M}$, a uniform weighting can still be maintained.

One might also criticize the use of uniform weighting by the fact that it diverges from the current approach taken in individual intelligence measurements. As we have explained in \ref{sec:indInf}, the universal measure of (individual) intelligence in equation \eqref{eq:univIndivIntell} -- as proposed by \citet{LH07}  -- weights the problems using a universal distribution. Following this approach, one would use $w_\mathcal {M}=2^{-K_U (j)}$ as the most appropriate weight. The measure of performance on the set of problems $\mathcal{M}$ is thus given by: $\bar{\Psi}(\mathcal {M},n|w_\mathcal {M}=2^{-K_U (j)}):=\sum_{j \in \mathcal {M}}2^{-K_U (j)} \bar{\Phi}(j,n)$ . An open question is now how one can aggregate this performance measure over several problem sets $\mathcal{M}$ in order to obtain a universal measure of collective intelligence $\bar{\Upsilon}$ . Another problem with this approach is that it is subject to the criticism we put already forward in \ref{sec:indInf}. The easiest problems receive the highest weight.

However, uniform weighting is still coherent with an approach based on algorithmic information theory. Again we use Kolmogorov Complexity for our collective intelligence measure. Yet, instead of expressing the complexity of the individual problems in $\mathcal{M}$, one can actually express the complexity of the whole set $\mathcal {M}$, which is in fact nothing else than the definition of our testing environment, previously denoted by $\mu$. The complexity $K_U (\mathcal {M})$ is simply the shortest input to the UTM $U$, which simulates all problems in $\mathcal{M}$. As mentions, the next step is to define a set $\mathcal{E}$ of different environments -- thus problems sets -- on which the group will be evaluated. Note then by $\bar{\Phi}(\mathcal {M},n|w_\mathcal {M}=\tfrac{1}{m})=\sum_{j \in \mathcal {M}}\tfrac{1}{m} \bar{\Phi}(j,n)$,  the performance measure using a uniform weighting. Then similar to the definition of universal  measure of individual intelligence in \eqref{eq:univIndivIntell}, one could express a measure of \textit{universal collective intelligence} as:
\begin{equation}
\bar{\Upsilon}(n)=\sum_{\mathcal {M}\in \mathcal{E}}2^{-K_U (\mathcal {M}) }  \bar{\Phi}(\mathcal {M},n|w_\mathcal {M}=\tfrac{1}{m})
\end{equation}

However, we are still not convinced of the use of a universal distribution. As a preferable measure of collective intelligence we propose for instance the average complexity of the testing environment on which the group performs significantly better than random (where the ``margin'' above random performance should be a parameter). 

As the space $\mathcal E$, and even the set of environments $\mathcal{M}$ with a specific complexity $K_U (\mathcal {M})=K$, is infinite, a difficulty of such an approach is however to develop a test which is executable in a finite time. Some ideas about how this might be done, can be found in \citet{HD10}.

\section{Use of the agents' ability in the voting process}
\label{sec:abilPub}

In the previous section we discussed how the group could use information about the problems' difficulties and whether this measure should actually be provided to the group. We will now do the same with the ability measure $\alpha$. This information might be used in two ways. First, it can be used in the voting process and second, in the allocation process. We will leave the latter point for future work and only discuss the first here.

In the voting process, $\alpha$ might be used to give more weight to the most intelligent agents. In \ref{ssec:aggregateVoting} we have defined three weighting systems: a majority voting system, a weighting system proportional to $\alpha$ and the optimal weighting system proportional to $\log⁡ \frac{P_\lambda(\alpha_i)}{1-P_\lambda(\alpha_i)}$. We have already made an extensive discussion about these weighting systems in \ref{sec:voteOneProb}.

\subsection{Should the group be provided with a measure of ability?}

There is however one question, which is still open: In how far are the agents able to find out about their ability themselves or must it be provided by the tester (assuming that it is actually known, which is not trivial)? Our answer to this question is similar to that of the previous point \ref{ssec:provideDiff} where we argued that the tester should not provide any information about the problems' difficulties. Similarly, we argue that the group should not be provided with an ability measure. And the reasons are actually identical to that of the previous point.

First of all, there exists (currently) no valid measure of intelligence. In the case of a monotonically increasing response function our parameter $\alpha$ can at least be used as a \textit{relative} measure of  ability: $\forall\alpha_l< \alpha_h  ,\forall\lambda:P_\lambda(\alpha_l)<P_\lambda(\alpha_h)$. Yet, any monotonically increasing transformation of $\alpha$ is also a valid measure. In most cases, the response function might however not even be a relative measure of ability due to phenomena such as agent specialization or different types of problems, in which case the response function is not monotonic anymore.

Second, the agents can find out -- while exploring the environment -- how well the agents perform on each of the problems. Again $\bar{\Psi}_{ij}$, the exponential moving average of agent $i$'s accuracy on problem $j$ contains the necessary information. This measure depends on both, the agent and the considered problem. Therefore, it takes into account any specific phenomenon we have just mentioned (specialization, problem types). As explained, $\bar{\Psi}_{ij}$ is an empirical measure of $P_{\lambda_j}(\alpha_i)$. And as we know, weights proportional to $\log⁡\tfrac{P_{\lambda_j}(\alpha_i}{1-P_{\lambda_j}(\alpha_i)}$ provide the optimal results. The group can thus simply use weights proportional to $\log \tfrac{⁡\bar{\Psi}_{ij}}{1-\bar{\Psi}_{ij}}$ in order to use the best weighting system possible.  

We have verified in figure \ref{fig:EexpMaPerf} that $\bar{\Psi}_{ij}$ varies indeed closely $P_{\lambda_j}(\alpha_i)$, without of course converging perfectly to it. In order to verify that the estimate $\bar{\Psi}_{ij}$ can successfully be used in the weighting scheme, let us consider a heterogeneous group of seven agents with abilities $\boldsymbol{\alpha}=\{1,2,3,4,5,6,7\}$ on a problem with difficulty\footnote{ The individual accuracies of the agents are thus 50.0\%, 50.7\%, 53.4\%, 57.6\%, 61.9\%, 65.9\% and 69.3\% respectively.} $\lambda=5$. Then the average group accuracy over 50 runs with 1000 rounds equals 73.1\% using the underlying value of the agents' accuracy $P_{\lambda_j}(\alpha_i)$. Using the empirical value $\bar{\Psi}_{ij}$, the accuracy drops only very slightly to 72.8\%. This is to compare to an average group accuracy of  68.1\% when majority voting is used. We conclude that $\bar{\Psi}_{ij}$ can indeed be used as a proxy for $P_{\lambda_j}(\alpha_i)$ in the weighting scheme.

Determining the right measure of agents ability to be used for voting is hence fairly easy and should also be manageable by the ``less intelligent'' groups. No indication must/should be provided by the tester.

Again, it should be noted that using an empirical value $\bar{\Psi}_{ij}$ is well suited for dynamical environments. Suppose that for one reason or another, the agent' accuracy changes over time. Then $\bar{\Psi}_{ij}$ will slowly take this into account and the voting weights are adapted accordingly.

\section{Imitating agents}
\label{sec:imitating}

Until now we have supposed that the agents provide their answer independently. This might be considered as a very restrictive hypothesis when studying ``collective'' intelligence. Typically one would expect that cooperation should result in positive interactions and increase the performance of the group.

We will now analyze a form of cooperation, which might possibly result in negative interactions. Up to now the agents' answers  have followed a binomial probability distribution, which implies that the answers  are independent from each other. This will change now. 

First, we will suppose that the $n$ agents make their decisions one after each other (sequentially), and not simultaneously as previously supposed. We will then suppose that the agents are somewhat ``lazy'' and inspire themselves from the answers provided previously. As before, the first agent will provide its (binary) response according to $r_1\sim P_\lambda(\alpha_1)$. The following agent will inspire itself partially from the previous one, so that:  $r_2\sim(1-\chi) P_\lambda(\alpha_2)+\chi r_1$, where $0<\chi<1$ is the \textit{imitation rate}. Similarly, the $l$th agent to decide provides an answer inspired by all previous answers:
\begin{equation}
r_l\sim(1-\chi) P_\lambda(\alpha_i)+\frac{\chi}{l-1} \sum_{i=1}^{l-1}r_i 
\end{equation}

One might expect that the resulting performance of the group is actually worse, even though -- or rather -- because the agents are actually collaborating/interacting. As explained in \ref{ssec:singleProbHomogen} for a homogeneous group of agents, the (independent) answers provided follow a binomial distribution $r_1,r_2,\hdots r_n\sim B(n,P_{\lambda}(\alpha))$. The average share of agents providing the correct answer will be the same as the accuracy of the agents $P_{\lambda}(\alpha)>50\%$.  Yet, the variance of this average share is $\sigma^2=\tfrac{P_{\lambda}(\alpha)\left(1-P_{\lambda}(\alpha)\right)}{n}$. Thus the higher the number of agents $n$, the higher is indeed the likelihood that the majority of agents votes for the right answer.

However, when we introduce imitating agents -- hence when the answers provided are not independent -- this is not given anymore. Take the extreme case where $\chi=1$. In this case the answer provided by all agents is that of the first agent. The average accuracy is the same as previously. However, its variance is that of the Bernoulli distribution (so the same as for one, here the first agent): $\sigma^2 =P_{\lambda}(\alpha)(1-P_{\lambda}(\alpha))$. Therefore, we cannot use the law of big numbers so as to increase the group's accuracy.

We have hence shown analytically that some specific form of collaboration/imitation will actually reduce the performance, at least in the particular case of homogeneous and fully imitating agents. Through our simulations we would like to generalize this result to other cases. For this we will work on one problem only. We will observe what happens when the agents are heterogeneous and not fully imitating $\chi<1$. It is also interesting to observe what happens if the abilities $\alpha_i$ are public knowledge. In this case, each agent could wait for the answer of the more able/intelligent agents and inspire itself from their answers. One might believe that such a setting will result in superior performance of the group, as the vote of the best agents receive a higher importance. Yet, given the previous explanations stating that the benefits of the law of big numbers are reduced, it is believed that this is actually detrimental to performance.

\subsection{Random imitation}

\begin{figure}[tb!]
\centering
\includegraphics[width=0.8\textwidth]{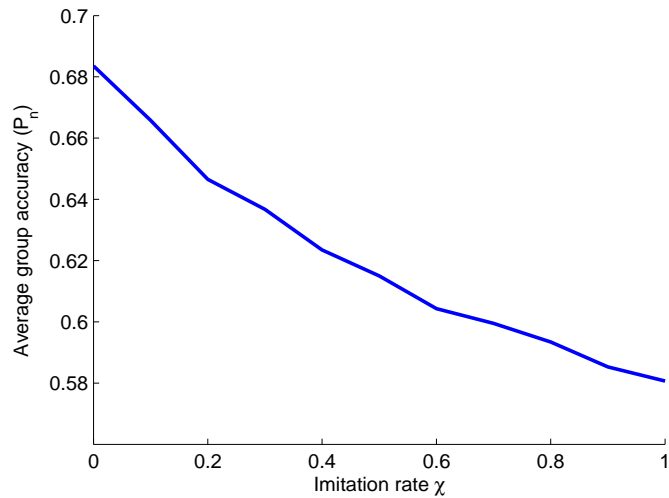}
\caption{ Average group accuracy for imitating agents providing their answer in a random order as a function of the imitation rate $\chi$. A group of seven agents with abilities $\boldsymbol{\alpha}=\{1,2,3,4,5,6,7\}$ works on a problem with difficulty $\lambda=5$}
\label{fig:Fimitating}
\end{figure}

In figure \ref{fig:Fimitating} we show the average group performance (over 30 runs of 1000 time steps) for imitating agents which provide their answers (hence imitate) in a random order.  The test setup is the same as in the previous section \ref{sec:abilPub}. That is, seven agents of ability $\boldsymbol{\alpha}=\{1,2,3,4,5,6,7\}$  work on a problem of difficulty $\lambda=5$\footnote{ The individual accuracies of the agents are thus 50.0\%, 50.7\%, 53.4\%, 57.6\%, 61.9\%, 65.9\% and 69.3\% respectively.}. A majority voting system is used. As can be observed, the performance decreases with the imitation rate. As explained this is due to the loss of independence of the individual votes, which increases the variance of the share of correct votes. When $\chi=1$, the answer provided by the group is the one provided by the first agent. As the latter is selected at random, the group's accuracy is the average of the individual accuracies which is 58.4\% in our case.

\subsection{Best imitation}

Let us now see what happens when the agents provide the answers in the order of their abilities; that is, from the most accurate to the least accurate. The agents are hence imitating the better agents. On the same setup as above, the performance increases, yet only slightly. Again, when $\chi=1$ the group's accuracy is that of the agent having provided its answer at first, which is here the best agent having an accuracy of 69.3\%. 

As explained above, there are actually two underlying effects at work here when the agents start imitating more and more. First, there is the increase of the answers' dependence, reducing the likelihood that a majority will vote for the right answer. Second, more importance is given to the decisions made by the best agents, which should increase the group's accuracy. Which of both effects is dominant depends actually on the specific setup which has been chosen. In our case, the accuracy of the best agent (69.3\%) is actually superior to the group using majority voting without imitation (68.1\%). It was thus to expect that the groups accuracy would increase when $\chi$ goes from 0 to 1. However, it is not necessarily -- actually rarely -- the case that the group's performance is lower than that of the best agent. It happened in our case as we are in presence of a very heterogeneous group with very good and very bad agents. When the bad agents do not imitate the good agents, their performance is very close to random and ``disturb'' the good agent (as discussed in \ref{ssec:lowPerfAgents}), which has a very negative impact on the group's performance. The latter might hence be increased when the agents start imitation the better agents. Imitation contributes hence positively here by increasing the weight of the good agents and to decrease that of the bad agents in a heterogeneous group.

\begin{figure}[tb!]
\centering
\begin{subfigure}[b]{0.8\textwidth}
        \centering
        \includegraphics[width=\textwidth]{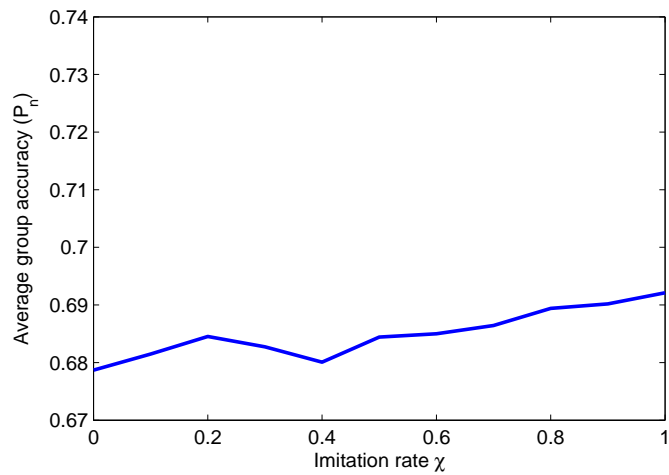}
        \caption{$\boldsymbol{\alpha}=\{1,2,3,4,5,6,7\}$}
\end{subfigure}
\begin{subfigure}[b]{0.8\textwidth}
        \centering
        \includegraphics[width=\textwidth]{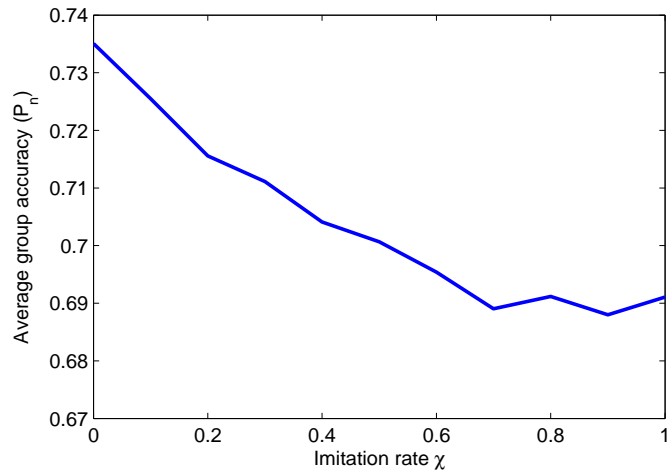}
        \caption{$\boldsymbol{\alpha}=\{4,4,4,4,5,6,7\}$}
        \label{sfig:FimitatingBest4444567}
\end{subfigure}
\caption{Average group accuracy for agents imitating the better agents as a function of the imitation rate $\chi$. A group of seven agents  works on a problem with difficulty $\lambda=5$.}
\label{fig:FimitatingBest}
\end{figure}

However, if the group is less heterogeneous -- i.e., the individual accuracies are more similar -- it is less important to increase the importance of the good agent and imitation is actually detrimental to performance. This is what can be observed in figure \ref{sfig:FimitatingBest4444567}, where we have replaced the bad (i.e. disturbing) agents by some better ones $\boldsymbol{\alpha}=\{4,4,4,4,5,6,7\}$\footnote{ The accuracies of the agents are hence now 57.6\% for the first four agents 61.9\%, 65.9\% and 69.3\% for the three remaining.}. In this case, imitation has a detrimental impact on the group's performance as the independence of the answers is more beneficial than having the agents imitating the answers of the better agents.

We can thus confirm that imitation, and also the imitation of the better agents, is most generally detrimental to performance. More precisely, this statement is true precisely for groups which are not excessively heterogeneous and/or sufficiently big so that the performance of the group with independent votes is higher than that of the best agent. Giving more importance to the most accurate agents should not be achieved by imitation, but rather by weighting. Actually, this means also that the independence of the agent's answers is not just a restrictive hypothesis of our report, but rather a beneficial property for the group's performance.

Our results are in coherence with those of others. \citet{Orl98} investigates the dynamics of a group where some agents are truly informed and others are pure imitators of the informed agents (their vote is determined through the majority vote of the informed agents). He shows that for a very small share of imitators, the latter's individual performance is actually superior to that of the individual informed agents. Also the imitators do not impact the collective performance negatively as they simply replicate/multiply the answers of the informed agents. However, he shows that when the share of imitators in the population is above a certain threshold, the individual performance of the imitators, but also that of the group decreases strongly as more agents start imitating.

\chapter{Analysis of the results}
\label{ch:sumResults}

\section{Modeling of intelligence}

Our work represents a step forward on how one can model (collective) intelligence. For this we used the approach defined by item response theory by stating a response function in equation \eqref{eq:respFunc}:
\begin{equation*}
P_{\lambda}(\alpha)=\frac{1}{2}+\frac{1}{1+\exp⁡\left(2\frac{\lambda}{\alpha}\right) } 
\end{equation*}

This approach was at the same time very simple, but still quite general. First of all, the model depends on two parameters only; one -- the ability $\alpha$ -- characterizing the agent and the other -- the difficulty $\lambda$ -- characterizing the problem. Moreover, this function is monotonically decreasing with $\lambda$ and monotonically increasing with $\alpha$. Also, it converges to an accuracy of 50\% as the problem gets more and more difficult.

All these properties however could easily be modified. In \ref{sec:singlePeak} we will see that with a slight modification in the previous function, one can also model single peaked functions. We argue however that the approach taken in \ref{ssec:specializedAgents} is actually even more general. In the latter section we defined -- instead of a constant ability for each agent -- an ability matrix $\alpha_{ij}$. Many specific situations could be modeled this way. We discussed the case where the agents are specialist in one problem \ref{sec:simpleSpecializedSim} or one specific type of problem (see \ref{sec:probTypes}). As mentioned, also non-monotonicity could be modeled this way, without even modifying the response function.   In \ref{sec:asymPerfWorseRand} we will briefly discuss agents with accuracies lower than random. 

\section{Collective intelligence tests}

Our approach has also allowed us to get some ideas about how collective intelligence test might be designed.

\subsection{How to introduce a social dimension into collective intelligence tests}

One innovative advancement of our approach deals with how to test social aspects of collective intelligence. We proposed that the group should not be tested on only one, but several problems at the same time. Therefore, the group must be able to organize itself so as to allocate its resources across all problems. As we could understand from our experiments, allocating the agents appropriately across the problems is not a trivial task. Each agent might for instance be good at a different problem. Allocating the agents in a way so that each agent is allocated to a problem at which he is good is not simple. Most important is however that the allocation task forces the group to communicate -- for instance about the agent's abilities and the problem's difficulties --, which is an important aspect of collective intelligence. 

\subsection{Dynamic environments}

We have also argued in \ref{ssec:dynEnvi} that dynamic environments -- i.e., environments in which for instance the problems' difficulties change over time -- are more difficult and might be exploited to discriminate groups. Two groups might perform similarly on static problems, yet one of them might be better in a dynamic environment than the other.

\subsection{Which information to provide?}

We have also provided an answer to which information should be provided to the group. We have argued that as few information as possible should be provided to the group. We have discussed here about whether  the problem's  difficulties (see \ref{sec:diffPub}) or the agent's abilities (see \ref{sec:abilPub}) should be provided to the group. None of these measures should be provided. This is because there exists no ``universal'' definition of the ability or the difficulty. Any measure of agent ability must be specific to a problem and any measure of problem difficulty must be specific to an agent. Moreover, the environment provides enough information about these measures. However, so as to exploit these measures successfully, the agents must be able to communicate about them. Hence letting the agents find out about their abilities and the problems difficulties and to exploit this information collectively, can actually be used as a part of the test. Again the importance of communication appears here, which is -- as mentioned -- an important aspect of collective intelligence.

\section{Collective decision making}

One of the most important issues in collective performance of a group is how the group makes a joint decision (for one problem or for several). We have observed several phenomena.

\subsection{The dynamics of odd and even number of agents}

We have explained on a theoretical basis that when one increases the number of agents on a problem, the performance of the group should increase. However, we observed that this increase is typically different when one goes from an even number of agents to an odd number than vice versa. 

One reason for this is that in a group with an even number of agents it might arise that the vote is undetermined and that a random flip must be used to decide about the group's vote. We have shown analytically that in a homogenous group using majority voting, an odd number of agents perform as well as the group with the nearest even number of agents.

\subsection{The importance of voting systems}

We have also shown the importance of the voting system. We have shown that in a majority voting system, the performance of good agents could significantly be hampered by the presence of bad agents. Yet, we know that there exists one and only one optimal voting system, in which each agent  whose answer is not random contributes positively to the group's performance. In fact, the optimal weighting system is proportional to $\log⁡\frac{P_{\lambda_j}(\alpha_i)}{1-P_{\lambda_j}(\alpha_i)}$. This holds also for agents whose performance is actually worse than random as we will discuss in \ref{sec:asymPerfWorseRand}. As we have shown in \ref{ssec:provideDiff}, all the ingredients of this system -- the individual accuracy on the problem $P_{\lambda_j}(\alpha_i)$ -- can be easily be estimated by the group. 

\subsection{Independence of votes}

In \ref{sec:imitating} we have analyzed what might happen if we drop the hypothesis of independent votes. We modeled this by supposing that the agents are imitating each other. This can be seen as a form of cooperation/communication between the agents. As we have shown, even when each agent actually imitates the better agent, one can typically (in cases of fairly homogenous and/or sufficiently big groups) expect that the performance of the group will  decrease as the agents' imitation rate increases. The independence of votes might hence be considered as a strength of the group.

Showing this has some interesting consequences for one of the most important collective intelligence systems: financial systems. The systems are supposed to be efficient as many agents make their decisions independently from each other. Each market participant makes errors, yet these errors tend to compensate by the fact that the average buying or selling decision of a big number of market participants leads to a price which reflect at each moment the intrinsic value of an asset. Nevertheless, one might criticize the hypothesis that the market participants act independently from each other. In fact, traders base their decisions frequently (not only on their own analysis, but also) on the trading behavior of others, mainly the most successful traders. Also, most traders base their decisions on statements made by big institutions such as rating agencies.  If the independence of market decisions is not given anymore, this efficiency mechanism goes out of force as we have shown.

This has already been advanced by others, yet in a less formal way. \citet{EdK10} talking about the over-reliance on rating agencies puts it like this:
\begin{quote}
``We know that a market system based on a multitude of decisions, with plenty of trials and errors, mistakes and successes, will produce a better allocation of capital than a centrally planned mechanism, where a few ``experts'' make all decisions. The latter may look attractive in theory but fails in practice. In fact experts routinely make bad decisions, because they can be wrong and they can be corrupted, and bad decisions applied globally then lead to massive misallocation of capital and dramatic failures.''
\end{quote}

Showing that in systems where agents inspire themselves from others are less performing -- also when the inspiration comes from the best agents -- has some important implications for the regulation of financial systems. Policy makers should take measures so as to ensure that market decisions become more independent. They should for instance make sure that trades can be made anonymously at the stock exchange and also that the involved amount of money is not revealed. Of course, the over-reliance on rating agencies must be reduced.

\section{Allocation models}

The task allocation problem is another important issue in collective intelligence.

Our approach has shown some interesting modification of the threshold allocation model which might also inspire other users of this model.

\subsection{Avoiding stimulus divergence}

We have proposed a feature avoiding the divergence of the stimulus. We included a factor $\zeta <1$ in the update rule in \eqref{eq:stimulusUpdateFull}:
\begin{equation*}
S_j\leftarrow\zeta S_j+\delta-\beta \frac{n_j}{n}-\beta' \bar{\Psi}_j
\end{equation*}

It is surprising that this simple, but effective feature has to our knowledge not been mentioned previously.

\subsection{Additional terms in the stimulus update rule}

We have also shown that the stimulus update rule can be adapted with some problem specific terms. In \eqref{eq:stimulusUpdateFull} we included a term $\bar{\Psi}_j$ reflecting the group's recent performance on the problem. Yet, we showed also that the term $\beta\tfrac{n_j}{n}$  should always remain in the model. This term can (in the standard model) be interpreted as the term forcing the group to allocate somewhat uniformly across the problems. Yet, complemented with other terms (and possibly in combination with a lower associated parameter) its purpose is rather to avoid extreme allocations where one (or more) problems are unserved and/or all agents are allocated to one problem.

In \ref{ssec:simDiffPub}, we showed also that it might be useful to transform the parameters of the update rule into functions of other parameters, here a measure of difficulty.

\subsection{Adaptation of the threshold update rule}

Finally our experiments suggest that including additional parameters in the threshold update rule might also improve the performance of our allocation model. In equation \eqref{eq:thresholdUpdateImproved}, we assured that the agents would typically be allocated to problems they are good at, by making the rule dependent on the agent's individual performance on the problems:
\begin{align*}
\theta_{ij}		&\leftarrow\theta_{ij}-\xi\cdot\bar{\Psi}_{ij}^2 \\
\theta_{ik}		&\leftarrow\theta_{ik}+\frac{\phi}{\bar{\Psi}_{ik}^2 }   \quad   \forall k \neq j
\end{align*}

\chapter{Discussions for future work}
\label{ch:futureWork}

In this chapter we will discuss further ideas for experiments, which can be performed in the context of the here explained approach.

\section{Use of agent ability in the resource allocation process}
\label{sec:mailman}

In \ref{sec:abilPub} we discussed the use of an ability measure in the voting process. This information might however also be used to allocate the agents among the problems. More precisely, we will have to allocate the agents to problems whose difficulty $\lambda$ are suited to the agent's level of intelligence $\alpha$. We will present here an idea about how this might be done. This means again that the difficulties $\lambda$ of the problems have to be public knowledge.

To do this, we can inspire ourselves from the mail retrieval problem. In the latter, a group of mailman has to retrieve the post from various cities. For this we use the typical threshold model to allocate the mailman to the cities. Yet, an additional piece of information is taken into account: the distance  $d_{z(i)j}$ between the current city $z(i)$ of mailman $i$, and other cities $j$ which might be served next. This information is important as the distance is proportional to the time the mailmen are occupied traveling from one city to another.

The probability for the mailman $i$ to serve city $j$ next should be higher for the nearer the city is. Following the approach of \citet{BSTD97}, we can incorporate the distance in our allocation model so that the probability of allocation becomes inversely proportional to it:
\begin{equation}
\mathrm{Prob}( i\mapsto j)=\frac{S_j^2}{S_j^2+\mu\theta_{ij}^2+\nu d_{z(i)j}^2}
\end{equation}

This means that each mailman will have a preferred zone of cities which will be served by him. Applied to our allocation problem, we first have to define for each agent a ``preferred zone'' of problems. So as to not waste our resources, the problem should neither be too easy (in which case the agent would better be allocated to a more difficult problem), nor too difficult (in which case the agent does not really contribute to solving the problem and should be allocated to an easier one). Keeping this in mind, it can be observed in figure \ref{fig:respFunc} that an intermediate level of accuracy is achieved when $\lambda_j \approx  \alpha_i$. We will therefore attempt to associate the problems in a way so as to respect this relation in the best way possible. We will use the ``distance function'': $d_{ij}=|\alpha_i-\lambda_j |$.

For the mail retrieval problem, the threshold updates are also modified. For a mailman $i$, the threshold is not only lowered for city $j$ he is currently serving, but also -- yet to a lesser extent --  for the neighboring cities $N(j)$:
\begin{align}
\theta_{ij} 	&\leftarrow\theta_{ij}  - \xi_0		\\ \nonumber
\theta_{in}  	&\leftarrow\theta_{in}  - \xi_1, \quad \forall n\in N(j), \ \xi_0  > \xi_1 		\\ \nonumber
\theta_{ik}		&\leftarrow\theta_{ik}  + \phi, \quad \ k\neq j,k\notin{N(j)}
\end{align}

This implies that the mailmen are more responsive to the stimuli -- i.e., the need to retrieve the post -- from a nearby city.

Applied to our allocation problems we want our agents to be more responsive to the stimuli -- i.e., the need to allocate more agents -- from problems with similar difficulties to the current problem. We could define two problems as being neighbors if their difficulties $\lambda$ do not differ by more than 10\%.

\section{Single peaked response functions}
\label{sec:singlePeak}

Until now we have supposed that the capability of correctly solving a problem (more precisely the corresponding probability $P_\lambda(\alpha)$) is a monotonically decreasing function of the difficulty $\lambda$ for each agent. Yet, for instance it might be that a very intelligent agent struggles solving very easy problems (say that they are ``too easy''). 

There are indeed real life examples of problems on which such a phenomenon arises. More precisely there are tests on which children perform better than adults. One such test is shown below in figure \ref{fig:Schoolbus}. The question is: in which direction is the bus driving? Most pre-school children answer this test correctly: The bus is driving to the left\footnote{The discussion is made for countries where traffic is right-hand. The answer is opposite for left-hand traffic countries, such as the United Kingdom or Australia.}. Most adults however are not able to provide the correct answer, also after a long period of consideration. The justification -- which is also provided by the pre-school children -- why the bus is heading leftwards is that one cannot see the entrance door which one could see if the bus would drive to the right.

\begin{figure}[tb!]
\centering
\includegraphics[width=0.6\textwidth]{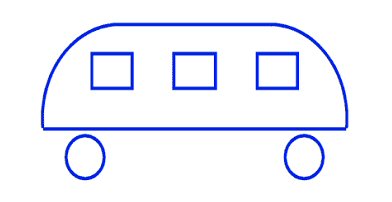}
\caption[In which direction is the bus driving?]{In which direction is the bus driving? \footnotemark}
\label{fig:Schoolbus}
\end{figure}

\footnotetext{Source: \url{www.sharpbrains.com/blog/2007/02/24/exercise-your-brains-visual-logic-brain-teaser}}

\begin{figure}[tb!]
\centering
\includegraphics[width=0.8\textwidth]{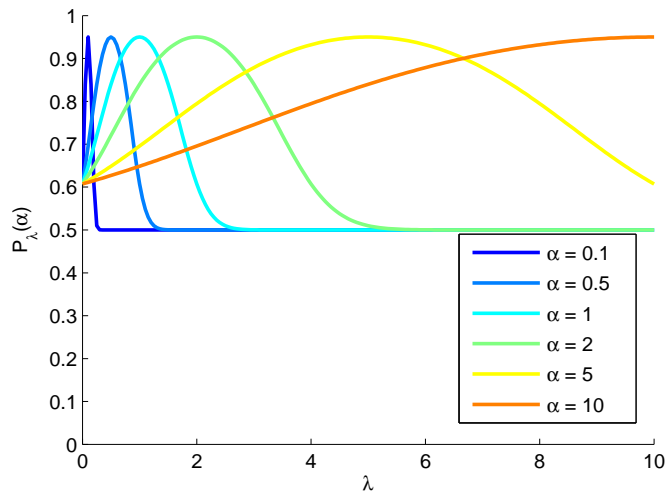}
\caption{Example of single peaked $P_{\lambda}(\alpha)$}
\label{fig:singlePeaked}
\end{figure}

It might therefore be interesting to analyze what happens if $P_{\lambda}(\alpha)$ is a single peaked function, i.e., each agent has a specific degree of difficulty $\lambda$ at which it performs best. An example of such a function might be for instance:
\begin{equation}
P_{\lambda}(\alpha)=\frac{1}{2}+\frac{0.9}{1+\exp⁡\left[2\left(\dfrac{\lambda-\alpha}{\alpha}\right)^2 \right]  }
\end{equation}
which is plotted in figure \ref{fig:singlePeaked}\footnote{$P_{\lambda}(\alpha)=\frac{1}{2}*\left[1+0.9*\exp⁡\left[2\left(\frac{\lambda-\alpha}{\alpha}\right)^2 \right]\right]$  yields similar results}.
 
Here, our measure of intelligence $\alpha$ plays two roles. First, the response function reaches its maximum for  $\lambda=\alpha$, and reflects thus in some sense the capability of solving difficult problems. Second, $\alpha$ reflects how big the variety of problems is the agent can solve as represented by the ``variance'' of the function $P_{\lambda}(\alpha)$. It is worth mentioning that $\alpha$ is now not even a relative measure of intelligence, even though it does still reflect the ``intelligence'' of the agents.

What we are actually doing here is in somewhat a continuous version of the agent-problem association/specialization we presented in \ref{ssec:specializedAgents}. For this we used a problem specific ability matrix $\alpha_{ij}$. Yet, defining an ability matrix is more general than modifying the response function $P_{\lambda}(\alpha)$. More precisely, by defining a problem specific ability matrix $\alpha_{ij}$ one can achieve -- for each couple agents-problem -- the same accuracies $P_{\alpha_{ij}} (\lambda_j)$ which could be achieved by modifying the response function $P_{\alpha_i} (\lambda_j)$ with constant abilities $\alpha_i$  for each agent. We have already discussed what happens when an agent is specialized to one specific problem in \ref{ssec:modelImprove}. We will therefore leave the setup explained here for future work.

Note that for the defined transfer function, a group using the allocation model inspired from the mailman defined in \ref{sec:mailman} would work very well, as this model tends to allocate the agents to problems in a way that $\alpha_i\approx  \lambda_j$.

\section{Different types of problems}
\label{sec:probTypes}

It would also be interesting to investigate what happens if we introduce several types of problems, $\pi$. Each agent $i$ has then not only one ability $\alpha_i$, but a vector of abilities $\alpha_i (\pi_j)$. Now we see that this is a generalization of the ability matrix $\alpha_{ij}$  we defined for the specialized agents in \ref{ssec:specializedAgents}. Hence, we will not discuss this further here and leave it over for future work.

\section{Intelligence affecting the allocation capability}

Until now we have assumed that the capability of solving the problem does not affect the task allocation algorithm. Yet, this is a strong hypothesis. It might very well be that very intelligent individuals are also better able to allocate themselves to the most appropriate problem. One might therefore test the impact of making the allocation algorithm dependent on the parameter $\alpha$. More precisely, we could add a random noise $\eta(\alpha_i)$ to the probabilities of switching to another problem as defined by the threshold model:
\begin{equation}
\mathrm{Prob}( i\mapsto j)=\frac{S_j^2}{S_j^2+\theta_{ij}^2}*\eta(\alpha_i)
\end{equation}

The size of this random noise ``perturbation'' is inversely proportional to the intelligence $\alpha_i$ of the agent. This reflects that less intelligent agents are less able to ``understand'' the task allocation algorithm and are hence less precise in determining the correct probabilities. As an example of the noise we might use a log-normal distribution with $\sigma =\tfrac{1}{\alpha}$ and $\mu=0$. The use of a log-normal distribution ensures that the probabilities remain positive. For high values of $\alpha$ the perturbation factor is close to one and the probabilities of the threshold model are barely modified. Yet, for the lower values of $\alpha$ the assignments become more random.

\section{Asymptotic performance worse that random}
\label{sec:asymPerfWorseRand}

Until now we have supposed that the function $P_{\lambda}(\alpha)$ tends asymptotically to 50\% -- the performance of a random classifier -- as the problems get more difficult. Yet, one might imagine a situation in which this probability becomes even worse than random (especially for humans). It might thus be interesting to analyze what happens when $P_{\lambda}(\alpha)$ becomes lower than 50\% for problems which are too difficult for the agent.

Typically one would expect that this decreases the group's performance when, for instance, majority voting is used. As we have discussed it in  \ref{sec:voteOneProb}, bad agents are able to disturb good agents. For agents having an accuracy below 50\% this is even more true.

It should however be noted that a group using the optimal weighting scheme is actually positively  affected by agents having an accuracy $P_{\lambda}(\alpha)<50\%$. In fact, their associated weights $\log\tfrac{⁡P_{\lambda_j}(\alpha_i}{1-P_{\lambda_j}(\alpha_i)}$ will be negative, which is actually equivalent to inverting their vote. 

\chapter{Conclusion}

In chapter \ref{ch:goal} we stated our goal as being to observe some interesting dynamics in collective intelligence and investigate how it might be tested. These goals have been met. Yet, our proposals for future work show that there is still much work to do.

What distinguishes our approach from that of others is that we investigate simultaneously the use of task allocation and joint decision making systems.

Our approach has brought advancements in several areas. 
First, we have brought ideas about how (collective) intelligence can be modeled. 
Second, we brought some ideas about how one can perform collective intelligence tests. Our simulations of such tests have provided us with some insights about their underlying dynamics. 
Then, we have also gained insights about collective intelligence, not just when tested, but more in general. 
Fourth, our proposals for future work might inspire some future analysis. 
Finally, for our experiments, we have refined the threshold allocation model and hence shown that it might be adapted to more complex situations.

The results of our approach can be used in several ways. First of all, they can be used by others who want to implement task allocation and/or joint decision making systems. Most importantly however, this piece of research should be understood as a first attempt to to develop collective intelligence tests. It complements the ongoing research for individual intelligence tests. We hope that this work will inspire further research in this area.

The big question which remains is whether what we are actually looking fore exists. Is it possible to measure (collective) intelligence and give a mathematical definition of it? Or is intelligence a vague concept such as ``consciousness' or even ``beauty'', which has been created by humans? It might very well be that every notion of intelligence must be specific to a particular (set of) problems.
And even if in the future a universal definition of intelligence might be found for individuals, this report shows that extending it to groups brings some considerable challenges.

\newpage

\phantomsection
\addcontentsline{toc}{part}{Bibliography \& Appendices}

\phantomsection
\addcontentsline{toc}{chapter}{Bibliography}
\bibliographystyle{plainnat}
\nocite{*}
\bibliography{mscth}

\newpage

\appendix
\chapter{Appendices}

\section{Proof of equation \eqref{eq:PEvenOdd}}
\label{sec:ProofPEvenOdd}

What we prove here is the following:
Accuracy of a group with an even number of agents is similar to that of the group with the nearest and lower odd number of agents:
\begin{equation} 
\nonumber
P_{2n}=P_{2n-1} \quad \forall n\in \mathbb{N}
\end{equation}

Let us start by an arbitrary even number of agents represented by $2n:n\in \mathbb{N}$:
\begin{equation}
P_{2n}=\sum_{k = \underbrace{\scriptstyle \lfloor 2n/2 \rfloor +1}_{=n+1}}^{2n} \binom{2n}{k} p^k q^{2n-k} +\frac{1}{2}\cdot\binom{2n}{n} p^n q^n
\end{equation}

Where the first term represents again the probability of all vote combinations which give a majority to the correct answer and the second term represents the probability that a coin flip is needed and successful. In the first term we will make a change to the index so as to start the sum in zero:$k'=k-n-1$. Also the factor in the second term can be simplified: 
\begin{equation}
\frac{1}{2}\cdot\binom{2n}{n}=\frac{1}{2}\cdot\frac{(2n)!}{n!\cdot n!}= \frac{1}{2}\cdot\frac{2n\cdot(2n-1)\cdots 1}{n(n-1)\cdots 1\cdot n!} = \frac{(2n-1)!}{(n-1)!\cdot n!} =\binom{2n-1}{n-1}
\end{equation}

Hence we get:
\begin{equation}
P_{2n}=\sum_{k'=0}^{n-1}\binom{2n}{k'+n+1} p^{k'+n+1} q^{n-k'-1} +\binom{2n-1}{n-1} p^n q^n
\end{equation}

We will now express the group accuracy for the corresponding odd number $2n-1$:
\begin{equation}
P_{2n-1}=\sum_{k=\underbrace{ \scriptstyle \lfloor (2n-1)/2 \rfloor +1}_{=n}}^{2n-1} \binom{2n-1}{k} p^k q^{2n-k-1}
\end{equation}

So as to let appear identical terms as in the expression of $P_{2n}$, we will apply a mathematical trick here. In our previous example we saw that a factor $p+q=1$ could be isolated in the accuracy expression of an even number of agents. Hence we will multiply $P_{2n-1}$ by the same factor:
\begin{equation}
(p+q)P_{2n-1}=\sum_{k=n}^{2n-1}\binom{2n-1}{k} p^{k+1} q^{2n-k-1} +\sum_{k=n}^{2n-1}\binom{2n-1}{k} p^k q^{2n-k}
\end{equation}

As before, we will modify the index of the sums. We would like the terms to have the same exponents. Hence, for the first sum we will apply $k'=k-n$ and for the second we will use $k'=k-n-1$:
\begin{eqnarray}
P_{2n-1}=\sum_{k'=0}^{n-1}\binom{2n-1}{k'+n} p^{k'+n+1} q^{n-k'-1} +\sum_{k'=0}^{n-2}\binom{2n-1}{k'+n+1} p^{k'+n+1} q^{n-k'-1}  \\
 +\binom{2n-1}{n-1} p^n q^n   \nonumber
\end{eqnarray}

The last term corresponds to the term with the negative index $k'=-1$ which appears in the second sum. Next we will use Pascal's rule $\binom{n}{k}=\binom{n-1}{k-1}+\binom{n-1}{k}$  to group the corresponding terms of both sums:
\begin{eqnarray}
P_{2n-1}=\binom{2n-1}{2n-1} p^2n+\sum_{k'=0}^{n-2}\binom{2n}{k'+n+1} p^{k'+n+1} q^{n-k'-1} \\
+\binom{2n-1}{n-1} p^n q^n \nonumber
\end{eqnarray}

The first term corresponds to the last term of the first sum ($k'=n-1$ ) having no equivalent in the second one. Yet, this term can again be re-injected into the sum:
\begin{equation}
P_{2n-1}=\sum_{k'=0}^{n-1}\binom{2n}{k'+n+1} p^{k'+n+1} q^{n-k'-1} +\binom{2n-1}{n-1} p^n q^n
\end{equation}

Which shows that $P_{2n}=P_{2n-1}$. What we have actually done, is to prove the formula \eqref{eq:PmajGeneral} known from multi-classifier systems.

\section{Description of the implementation}

So as to perform the previously described experiments, a program has been implemented in Java. We will very briefly describe this implementation here. We will start by describing the involved classes. Thereafter the main idea of the code will be explained. 

	\subsection{Classes}

Our implementation contains four classes:

\subsubsection{Agent}

This class represents the agents. One important attribute is for instance the ability $\alpha$ of the agent. The main role of this class is to implement the functions necessary to solve the allocation problem using the threshold model. So as to specify the behavior of the agents, several \textit{enum} classes are used. One specifies for instance the joint decision taking system which is used by the agents: Absolute voting or weighted voting, either by their $\alpha$ or using the optimal weighting scheme $\log⁡\tfrac{P_{\lambda}(\alpha)}{1-P_{\lambda}(\alpha)}$. Another specifies which probabilistic rule for problem $\mathrm{Prob}( i\mapsto j)$  is used. This can either be the standard version, the version with a log normal noise depending on $\alpha$ or the one stemming from the mailman problem. Finally, we also need to specify which type of response function is used. There is the standard, monotonically decreasing and a specialized, single peaked version. Also there are two versions where the agents are specialized to a problem, one with a simplified, static threshold update rule and another with the standard update rule. Finally, there are also two versions of lazy agents, one where the agents imitate the better agents and another where the order of decision taking is randomized.

\subsubsection{Problem}

This class represents the problems. An important attribute is for instance the difficulty $\lambda$ of the agent. The main role of this class is to implement the functions necessary to evaluate the agents on the problem they have chosen. Again, problems have distinct behaviors, depending on how the stimulus update rule is implemented. Here we distinguish between a version with and without a term related to the share of agents specialized to the problem. Also we distinguish a version where the parameters of the update rule depend on the difficulty if we decide that this information is known to the agents.

\subsubsection{Test}

This class represents the test, in which the agents have to face various problems. It has been implemented using a singleton pattern, i.e., there exists only one instance of this class. It implements the principal routine presented below.

\subsubsection{Run}

This class implements the main function. So as to run the implementation, one parameter has to be provided: the name of an XML file responsible for importing the specifications of the run. An example of such a specification file can be found in figure \ref{fig:spec}.

\lstset{language=XML, frame=single, numbers=left, numberstyle=\footnotesize , tabsize=2, 
	basicstyle=\footnotesize ,
	keywordstyle=\color[rgb]{0,0,1},
	commentstyle=\color[rgb]{0.133,0.545,0.133},
	stringstyle=\color[rgb]{0.627,0.126,0.941},
	morekeywords={encoding, settings, agent,pSwitch,xi,phi,thetaMin,thetaMax,problem,delta,
	betaAvgAccuracy,betaNbAgents,lambda,agents,alpha,test,nbRounds}
}

\begin{figure}[h!]
\begin{adjustwidth}{-20pt}{-20pt}
\centering
\end{adjustwidth}
\caption{Example of an XML specification file}
\label{fig:spec}
\end{figure}

\subsection{Principal routine}

The execution of the implementation is divided into five steps, of which the last four are executed in a loop:
\begin{description}
\item[Initialization] During the initialization phase, all instances of the classes will be initialized as specified in the XML file.
\item[Loop]\hfill
\begin{description}
\item[Switch problems] The agents select a new problem according to the threshold model.
\item[Evaluate] All agents have to provide their answer to the problem they have chosen and to vote the group's answer.
\item[Update] All agents update their thresholds $\theta_{ij}$ and all problems update their stimuli $S_j$.
\item[Report] So as to be able to analyze the results of our experimental phase, after each round the classes \textit{Agent} and \textit{Problem} will both print some descriptive statistics about their state into a text file. Both files can then be imported into Matlab so as to analyze them. In Matlab we dispose also of method to launch the implementation repeatedly under various specifications. Thisis very useful for our experiments, for instance in order to make comparisons between different specifications and to obtain statistically significant results.
\end{description}
\end{description}

The Java environment of Matlab provides also the necessary functions to make automated modifications of the parameters specified in the XML file.

\end{document}